\documentclass{article}

\usepackage[main, final]{neurips_2026}

\usepackage[utf8]{inputenc} 
\usepackage[T1]{fontenc}    
\usepackage{hyperref}       
\usepackage{url}            
\usepackage{booktabs}       
\usepackage{amsfonts}       
\usepackage{nicefrac}       
\usepackage{microtype}      
\usepackage{xcolor}         

\usepackage[utf8]{inputenc}
\usepackage[T1]{fontenc}
\usepackage{hyperref}
\usepackage{url}
\usepackage{booktabs}
\usepackage{amsfonts}
\usepackage{amsmath}
\usepackage{amssymb}
\usepackage{nicefrac}
\usepackage{microtype}
\usepackage{xcolor}
\usepackage{graphicx}
\usepackage{subcaption}
\usepackage{multirow}
\usepackage{tabularx}
\usepackage{array}
\usepackage{enumitem}
\usepackage{bbding}       
\usepackage{pifont}       
\usepackage{algorithm}
\usepackage{algorithmic}
\usepackage{wrapfig}
\usepackage{tikz}
\usetikzlibrary{arrows.meta, positioning, shapes.geometric, fit, backgrounds}

\definecolor{mydarkblue}{rgb}{0,0.08,0.45}
\definecolor{myfavblue}{rgb}{0.1176, 0.392, 1.0}
\usepackage{nicefrac}

\usepackage{listings}

\definecolor{dkgreen}{rgb}{0,0.6,0}
\definecolor{gray}{rgb}{0.5,0.5,0.5}
\definecolor{mauve}{rgb}{0.58,0,0.82}
\definecolor{lightgray}{HTML}{DDDDDD}
\usepackage{amssymb}
\usepackage{pifont}
  \definecolor{orange}{HTML}{ff7f0e}
  \definecolor{blue}{HTML}{1f77b4}

\usetikzlibrary{calc}
\usepackage{longtable}

\usepackage{amsmath}
\newcommand{\repeatcommand}[2]{%
  \ifnum#1>0
    #2%
    \repeatcommand{\numexpr#1-1\relax}{#2}%
  \fi
}

\hypersetup{
  colorlinks,
  linkcolor={red!50!black},
  citecolor={green!60!black},
  urlcolor={blue!80!black}
  }
\title{Seeing Together:\\Multi-Robot Cooperative Egocentric Spatial Reasoning with Multimodal Large Language Models}

\author{%
Kunyu Peng$^{1,2,}$\thanks{Correspondence: kunyu.peng@kit.edu and kailun.yang@hnu.edu.cn} \quad 
Zhikun Zhou$^3$ \quad 
Kailun Yang$^{3,*}$ \quad 
Di Wen$^{1}$ \quad 
Ruiping Liu$^1$ \\
\textbf{Yufan Chen$^{1,4}$} \quad 
\textbf{Junwei Zheng$^{1,6}$} \quad
\textbf{Hao Shi$^{5,7}$} \quad 
\textbf{Yi Zhou$^3$} \\ 
\textbf{M. Saquib Sarfraz$^{8}$} \quad
\textbf{Danda Pani Paudel$^2$} \quad
\textbf{Luc Van Gool$^2$}
\\
$^1$Karlsruhe Institute of Technology 
$^{2}$INSAIT, Sofia University ``St. Kliment Ohridski''
\\
$^3$Hunan University
$^4$University of Oxford
$^5$Zhejiang University \\
$^6$ETH Zurich
$^7$Ant Group
$^8$Mercedes-Benz Tech Innovation
}

\begin{document}

\maketitle

\begin{abstract}
Multimodal Large Language Models (MLLMs) have made substantial progress in egocentric video understanding, but their ability to reason cooperatively from multiple embodied viewpoints remains largely unexplored. We study this problem through \emph{multi-robot cooperative dynamic spatial reasoning}, where a model must answer spatial, temporal, visibility, and coordination questions by integrating synchronized egocentric videos from a team of moving robots. To support this setting, we introduce \textbf{CoopSR}, the first benchmark for this task, together with \textbf{EgoTeam}, a multi-robot egocentric QA dataset. EgoTeam contains $114,227$ QA pairs spanning $19$ question types, four difficulty tiers, and three team sizes in Habitat and iGibson, along with a real-world test set of around $2,326$ QAs collected using two quadruped robots. We further propose \textbf{SG-CoR} (\emph{Spectral and Geometry-Aware Cooperative Reasoner}), an MLLM framework for fine-grained cooperative spatial reasoning. SG-CoR combines dynamics-aware multi-robot frame sampling, spectral- and geometric-guided view fusion, and geometry- and motion-informed prompt distillation, enabling the model to benefit from privileged robot-pose supervision during training while requiring only egocentric videos at test time. Across $22$ MLLM baselines, SG-CoR consistently improves cooperative reasoning, outperforming the strongest fine-tuned baseline by \textbf{+3.87\%} on Habitat and \textbf{+7.12\%} on iGibson. It also shows stronger generalization to unseen team sizes and real-world robot tests. Code can be found at \url{https://github.com/KPeng9510/seeing-together.git}.
\end{abstract}

\section{Introduction}

Embodied AI is rapidly shifting from isolated agents to teams of robots operating in shared physical spaces, including warehouse fleets, search-and-rescue squads, inspection drones, and household assistants~\cite{egolife,ego4d,egoplan,kim2026ma}. 
In these settings, no single robot has a complete view of the world. Each agent observes only a partial, egocentric, and often noisy slice of the scene. Answering even a seemingly simple operator query, such as ``Which robot is pushing the object?'' or ``Can any robot see the blocked doorway?'', may require integrating observations across robots, grounding objects across viewpoints, reasoning about inter-robot spatial relations, and building a team-level understanding of the scene. Beyond perception alone, real-world deployment also requires robots to coordinate tightly during physically coupled tasks, such as two robots jointly carrying a long object through cluttered spaces, where failures in shared spatial understanding can lead to collisions, blockage, or unsafe interactions. We call this capability \emph{cooperative spatial reasoning}: reasoning over multi-robot egocentric video streams to build a shared understanding of a dynamic environment.

Despite its importance, this setting remains largely unexplored in current Multimodal Large Language Models (MLLMs). 
Existing MLLMs~\cite{zhang2025videollama,egolife,zhang2024llavanextvideo,hui2024qwen2} are primarily trained and evaluated on single-view images or videos, often from a single human or robot perspective. They typically treat visual tokens as a flat sequence and lack mechanisms for explicitly aligning, filtering, and fusing observations from multiple embodied agents. A naive extension—simply concatenating $N$ robot video streams—is not scalable: the number of visual tokens grows rapidly with team size, while adjacent frames and overlapping viewpoints introduce substantial redundancy.
More importantly, concatenation alone does not teach the model \emph{how} to reason collaboratively across agents, such as determining which robot has the best viewpoint, resolving contradictory observations, or inferring scene-level structure from distributed egocentric evidence.

Existing benchmarks~\cite{fan2019egovqa,jia2022egotaskqa,EgoMe} do not adequately address this problem. Egocentric VQA benchmarks~\cite{EgoMe,cheng2024egothink} mainly consider a single observer, while EgoExo QA benchmarks~\cite{jung2025egoexo,he2025egoexobench} focus on reasoning between a fixed exocentric view and one moving egocentric view.
Recent VQA benchmarks for moving egocentric agents, \textit{e.g.}, MA-EgoQA~\cite{kim2026ma}, assume that the agent already possesses sufficient spatial reasoning ability in dynamic scenes and primarily emphasize lifelong human behavior understanding, rather than fine-grained joint spatial reasoning, robot action understanding, and environment awareness.
In contrast, our benchmark is uniquely designed to evaluate whether MLLMs can understand a shared dynamic environment through the eyes of a robot team. It concentrates on cross-view grounding, inter-robot spatial relations, team-level scene understanding, and fills a critical gap by directly assessing team-level egocentric perception and reasoning for multi-robot embodied intelligence.

To fill this gap, we introduce \textbf{CoopSR}, a benchmark for \emph{cooperative multi-robot visual question answering}, and \textbf{EgoTeam} dataset collected from both simulators and real-world labs.
\textbf{EgoTeam} contains $114K$ QA pairs covering $19$ QA types along five axes: spatial, exploration, relational, robot action, and visibility reasoning. The questions span four levels, from egocentric scene understanding to cooperative scene perception, and three team sizes in habitat, iGibson, and real-world environments.
\textbf{CoopSR} evaluates $22$ MLLMs on synchronized egocentric observations from variable-size robot teams, targeting key skills such as spatial awareness, robot-object relations, scene composition, temporal robot action reasoning, mutual visibility, shared-object perception, team belief updates, and counterfactual coordination. Unlike prior VQA benchmarks focused on single-agent perception, \textbf{CoopSR} evaluates whether models can integrate partial multi-robot viewpoints, reason about complementary observations and inter-robot relations, and infer coherent scene-level answers from distributed embodied evidence.

\begin{figure}[t!]
\centering
\includegraphics[width=0.9\linewidth]{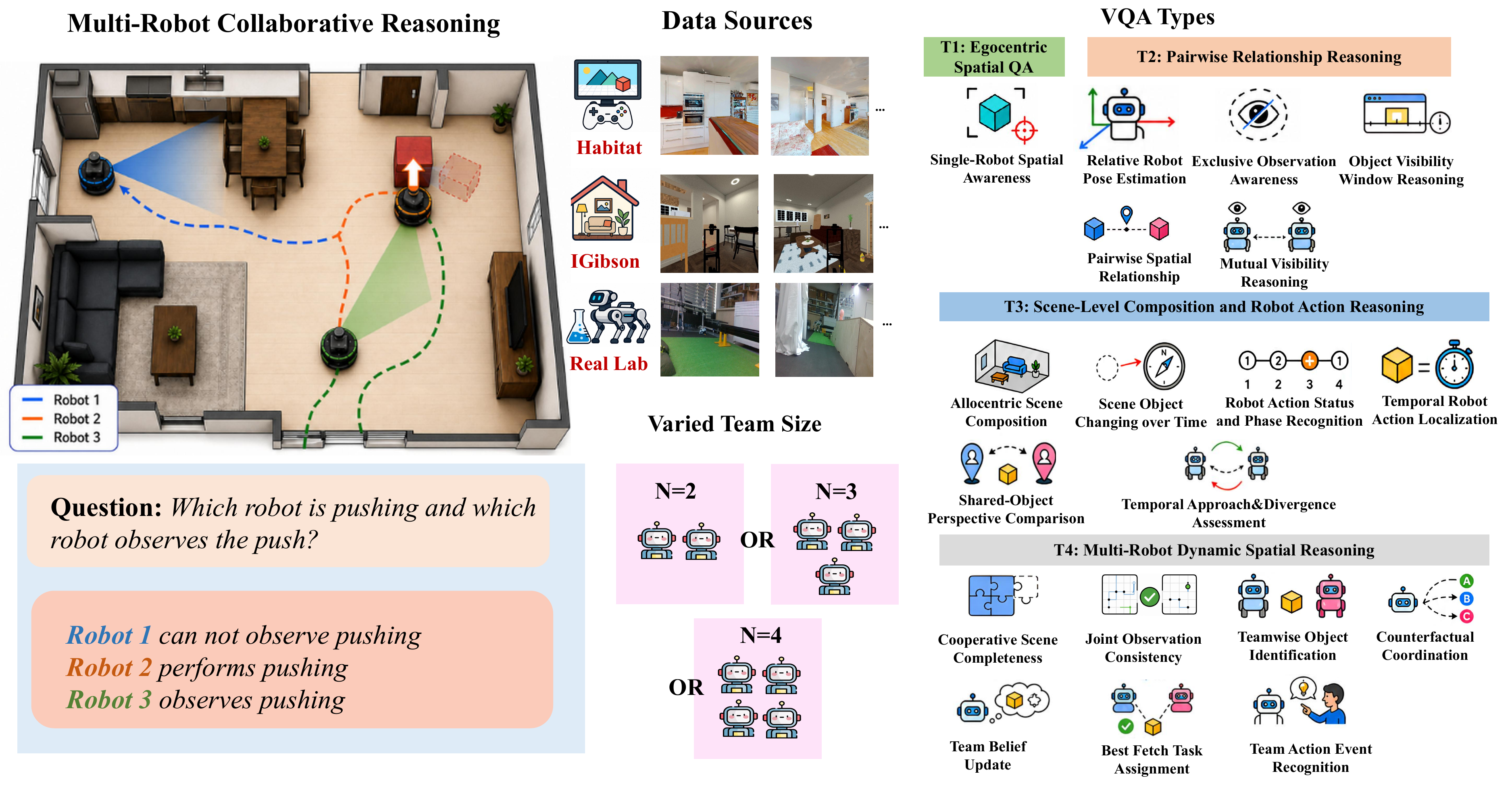}
\vskip-1ex
\caption{Overview of the \textbf{CoopSR} benchmark.
It evaluates spatial, temporal, visibility, and coordination reasoning using synchronized egocentric views from variable-size robot teams in simulated and real-world environments.}
\vskip-5ex
\label{fig:teaser}
\end{figure}

However, existing models~\cite{jeong2025videorag,zhang2025videollama,zhang2024llavanextvideo,hui2024qwen2,bai2025qwen3vl,wang2025internvl3} perform reasonably well on single-view reasoning~\cite{cheng2024egothink,EgoMe} or EgoExo setting~\cite{huang2024egoexolearn} but degrade sharply on questions requiring inter-robot integration, cross-team-size generalization, or conflicting viewpoints. 
This failure pattern suggests a structural bottleneck: current MLLMs lack inductive biases for agent identity, geometric co-visibility, and multi-robot egocentric view fusion.
Motivated by these findings, we propose \textbf{SG-CoR} (\emph{Spectral and Geometry-Informed Cooperative Reasoner}), an MLLM architecture designed for cooperative spatial reasoning. 
\textbf{SG-CoR} has three key components. 
First, a training-free two-stage sampler compresses each robot stream by jointly selecting the most query-relevant frames and refining them using temporal frequency energy ranking over visual features. 
Second, a spectral and geometry-informed attentional block fuses robot observations while injecting a geometry-prior from the data-rich simulator and the spectral cues into attentional fusion. 
Ground-truth poses are used only during training, and deployment requires no external localization oracle. 
Third, a pose-aware teacher distills physical priors into the student's prefix tokens, aligning the language prompt with the pose prior provided by the simulator. The student prompt is utilized as geometric prior during deployment, as the real world cannot be guaranteed to provide precise pose cues. 

On the \textbf{CoopSR} benchmark, \textbf{SG-CoR} consistently outperforms zero-shot, SFT, RAG~\cite{kim2026ma,jeong2025videorag}, and key frame selection~\cite{song2026ktv,tang2026tspo} baselines across different evaluation settings. The best \textbf{SG-CoR} variant, using a Qwen2.5-VL~\cite{bai2025qwen25vl} backbone, achieves an average accuracy of $70.55\%$ on Habitat~\cite{puig2023habitat} and $70.82\%$ on iGibson~\cite{li2021igibson}, improving over the strongest baseline by $+3.87\%$ and $+7.12\%$, respectively. \textbf{SG-CoR} also shows superior cross-MLLM-backbone generalizability.
These results indicate that \textbf{SG-CoR} benefits from an explicit cooperative reasoning design. 
Further evaluation on cross-team-size and sim-to-real generalizability further demonstrates the superiority of our proposed method.
In summary, our contributions are as follows:
\begin{itemize}[leftmargin=*, itemsep=1pt]
    \item We introduce \textbf{cooperative spatial reasoning} as a new embodied AI problem, where an MLLM must answer language queries by integrating synchronized egocentric observations from multiple moving robots.

    \item We present \textbf{EgoTeam}, a dataset for cooperative multi-robot visual question answering, containing $>114K$ question-answer pairs across $19$ QA types in simulated and real-world environments. We conduct a comprehensive evaluation of $22$ MLLM baselines on this dataset and build this first \textbf{CoopSR} benchmark.

    \item We propose \textbf{SG-CoR}, a cooperative MLLM framework that introduces a spectral energy-aware multi-robot relevant frame sampler, spectral and geometry- and motion-informed multi-robot fusion, and geometry- and motion-informed prompt-space distillation for scalable reasoning over multi-robot egocentric videos without requiring pose data during test time.
    \textbf{SG-CoR} delivers state-of-the-art performance across Habitat, iGibson, cross-team-size, and sim-to-real settings.
\end{itemize}
\section{Related Work}
\label{sec:related}

\noindent\textbf{Multi-agent perception and cooperation.}
Existing works on multiple embodied agents study cooperation across diverse environments~\cite{long2024teamcraft,wu2026multiworld,zhao2025empowering,chai2025causalmace,qin2025robofactory,zhang2024combo,savva2026solaris}, with systems such as CoELA~\cite{CoELA}, Co-NavGPT~\cite{co-navgpt}, OmniVLN~\cite{liu2026omnivln}, DeCoNav~\cite{zhou2026deconav}, AirCopBench~\cite{zha2026aircopbench}, TZPP~\cite{wang2026can_walk_robotic_dog}, COHERENT~\cite{liu2025coherent}, and CoMaTrack~\cite{liu2026comatrack} using LLMs/VLMs for multi-agent planning, coordination, and exploration. 
Other studies address more realistic constraints such as limited communication and partial observability~\cite{ace, furnmove}, while structured or leadership-based prompting has been shown to improve teamwork efficiency~\cite{guo2024embodied}. 
PARTNR~\cite{partnr} further provides a large-scale human-robot collaboration benchmark.

\noindent\textbf{Multimodal large language models.}
The past two years have seen rapid progress in MLLMs that combine vision encoders with LLM backbones: Video-LLaVA~\cite{lin2024video}, InternVL~\cite{chen2024internvl}, Qwen2.5-VL~\cite{bai2025qwen25vl}, Qwen3-VL~\cite{bai2025qwen3vl}, LLaVA-Next-Video~\cite{zhang2024llavanextvideo}, and VideoLLaMA3~\cite{zhang2025videollama}.
These models are trained on single-image or short-video data and treat all visual tokens uniformly, with no mechanism for agent-level identity or inter-agent geometric constraints.
We systematically evaluated $22$ MLLMs, including zero-shot MLLMs, supervised fine-tuned MLLMs, Retrieval-Augmented-Generation methods~\cite{kim2026ma,jeong2025videorag}, and key frame selection methods~\cite{tang2026tspo,song2026ktv}. 
Existing MLLMs lack cross-robot alignment, multi-view fusion, and team-level belief modeling, therefore, we propose \textbf{SG-CoR}, which integrates dynamics-aware sampling, geometry-guided fusion, and geometry- and motion-informed prompt distillation for cooperative spatial reasoning.

\noindent\textbf{Benchmarks of egocentric video question answering.}
Early egocentric vision research~\cite{de2009guide, fathi2012social, lee2012discovering, ryoo2013first, bambach2015lending, mueller2017real, palazzi2018predicting, liu2024coarse} was driven by foundational but relatively small datasets such as ADL~\cite{ADL}. 
The field later expanded through large-scale benchmarks including EPIC-KITCHENS~\cite{epickitchen}, Ego4D~\cite{ego4d}, and EPFL-Smart-Kitchen~\cite{bonnetto2025epfl}, which broadened egocentric video understanding to diverse daily activities.
Subsequent datasets targeted more specialized settings, including procedure learning~\cite{EgoProceLECCV2022, schoonbeek2024industreal,zhu2026egosound}, collaborative assistance~\cite{wang2023holoassist}, and joint egocentric-exocentric understanding~\cite{egoexo, huang2024egoexolearn}. 
Building on these resources, recent benchmarks have advanced temporal reasoning, forecasting, and planning in first-person video~\cite{lin2022egocentric, darkhalil2022epic, plizzari2022e2, tokmakov2023breaking, mangalam2023egoschema, li2024seed}.
Zhang~\textit{et al.}~\cite{zhang2025egonight} propose pairwise egocentric video question answering in day and night time.
More recent efforts further explore long-term video understanding~\cite{ye2024mm, chandrasegaran2024hourvideo1hourvideolanguageunderstanding}.
Jung~\textit{et al.}~\cite{jung2025egoexo} and He \textit{et al.}~\cite{he2025egoexobench} focus on egocentric and exocentric videos for cross-view video understanding.
MA-EgoQA~\cite{kim2026ma} focuses on long-term human behavior reasoning using multiple agents, which focuses on coarse-level QAs. 
Unlike these benchmarks, our  \textbf{CoopSR} directly targets fine-grained cooperative spatial reasoning from synchronized multi-robot egocentric videos. It evaluates whether MLLMs can integrate partial and dynamic robot viewpoints to reason about inter-robot relations, shared scene structure, visibility, and team-level coordination.

\section{CoopSR-Bench: Benchmark and Datasets}
\label{sec:bench}

\begin{table*}[!t]
\centering
\caption{Comparison of our \textbf{EgoTeam} with other egocentric QA datasets.}
\vskip-1ex
\label{tab:compare_benchmark}
\resizebox{\textwidth}{!}{
\begin{tabular}{@{}llccccccc@{}}
\toprule
\centering \textbf{Dataset} & \textbf{Source}               &\textbf{\#QAs} & \textbf{Size (hrs)} & \textbf{\#Clips} & \textbf{Dur./Clip} & \textbf{MultiView} & \textbf{Inter Robot Dynamic } & \textbf{Robot Pose}\\
\midrule

EgoSchema~\cite{mangalam2023egoschema}   &    Ego4D~\cite{ego4d}             &    5K     &250        &5,063         &3 min           &  \textcolor{red}{\ding{55}}    &          \textcolor{red}{\ding{55}}     & \textcolor{red}{\ding{55}}           \\
EgoPlan-Bench~\cite{egoplan}   &    Ego4D~\cite{ego4d} \& EPIC-KITCHENS~\cite{epickitchen}                &   4.9K     & -        &4,939         & -            & \textcolor{red}{\ding{55}}       &          \textcolor{red}{\ding{55}}  & \textcolor{red}{\ding{55}}   \\
EgoThink~\cite{cheng2024egothink}   &    Ego4D~\cite{ego4d}                &    0.7K     &  -       & 595         & -           &     \textcolor{red}{\ding{55}}      &          \textcolor{red}{\ding{55}}    & \textcolor{red}{\ding{55}}     \\
EgoMemoria~\cite{ye2024mm}    &      Ego4D~\cite{ego4d}                    &       7K   &  -    &     629   &     30 s to 1 h     &     \textcolor{red}{\ding{55}}       &          \textcolor{red}{\ding{55}}  & \textcolor{red}{\ding{55}}   \\
HourVideo~\cite{chandrasegaran2024hourvideo1hourvideolanguageunderstanding}     &   Ego4D~\cite{ego4d}         &     12.9K    & 381        &   500      &     20 min to 2 h     &    \textcolor{red}{\ding{55}}     &          \textcolor{red}{\ding{55}} & \textcolor{red}{\ding{55}}   \\ 
EgoLifeQA~\cite{egolife}     &  EgoLife~\cite{egolife}         &   3K    & 266        &   6      &      44.3 h    &    \textcolor{teal}{\checkmark} & \textcolor{red}{\ding{55}}    & \textcolor{red}{\ding{55}}            \\
\midrule
\textbf{EgoTeam (ours)}     &  - &   114K   & 385.7        &   21,124      &    1 min to 6 min   &    \textcolor{teal}{\checkmark}    &    \textcolor{teal}{\checkmark}    &    \textcolor{teal}{\checkmark}   \\
\bottomrule
\end{tabular}}
\vskip-3ex
\end{table*}

\noindent\textbf{Data collection and annotation.}
\noindent\underline{Metadata collection.} 
We first collect multi-robot egocentric data in Habitat~\cite{puig2023habitat} and iGibson~\cite{li2021igibson}. 
For each scene, we deploy teams of $2$, $3$, and $4$ robots and generate $15$ exploration episodes with diverse trajectories and robot-object interaction events, such as object pushing, yielding diverse layouts, viewpoints, and temporal dynamics.
For each team size $N \in \{2,3,4\}$, episodes are collected
in separate simulator runs under a common collection protocol
and comparable scene distributions. Robot trajectories are
generated jointly for each team and are not paired across
team sizes.
For each episode, we record robot poses, pairwise relative poses, egocentric RGB-D videos, semantic segmentation videos, and object to object/robot spatial relationships per timestep. 
These annotations support QA construction over egocentric perception, cross-view association, temporal reasoning, and multi-robot coordination. 
We also collect a real-world test set by exploring a room with two quadruped robots and using motion capture systems to provide pose metadata for QA creation; details are provided in the appendix.

\noindent\underline{QA annotations.}
For iGibson~\cite{li2021igibson} and Habitat~\cite{puig2023habitat}, we use simulator-recorded scene attributes as metadata for QA generation and then rephrase the QAs by GPT-4o~\cite{openai2024gpt4o}.
QA pairs are derived from the aforementioned extracted metadata.
This enables scalable QA generation while ensuring that each question is grounded in the scene state and has an unambiguous metadata-derived answer.
For the real-world quadruped robot dataset, we manually annotate scene captions as metadata at adaptive intervals of fewer than $30$ frames to avoid obvious scene changes within each interval, and then create QAs based on metadata. 
In total, we collect $1.75h$ of real-world data across $13$ scene settings. Additional statistics and comparisons with existing egocentric datasets are shown in Table~\ref{tab:compare_benchmark}. 
To improve evaluation reliability, four annotators manually review QA pairs in the validation and test splits to ensure correctness with a majority agreement ratio of $0.9$.

\noindent\textbf{QA taxonomy.}
Our QA taxonomy evaluates multi-robot egocentric reasoning across four hierarchical reasoning tiers of increasing cooperative scope, ranging from local spatial perception to team-level coordination, as illustrated in Figure~\ref{fig:teaser}.

\noindent\underline{T1: Egocentric spatial QA.}
T1 level evaluates spatial understanding from a single robot's first-person observation. It includes single-robot spatial awareness, such as recognizing object locations, directions, distances, and layouts.

\noindent\underline{T2: Pairwise relationship Reasoning.}
T2 level focuses on relationships between two entities, including robots, objects, and viewpoints. It evaluates whether a model can identify information visible to only one robot, reason about object visibility and occlusion over time, determine pairwise spatial relations such as left/right, front/behind, near/far, and facing direction, and assess whether two robots can see each other or share overlapping fields of view.

\noindent\underline{T3: Scene-level composition and robot action reasoning.}
T3 level requires integrating multiple egocentric observations into a coherent scene-level representation. It covers allocentric scene reconstruction, temporal tracking of object changes, recognition of robot action phases, estimation of action duration, comparison of shared objects across viewpoints, and reasoning about whether robots approach or move away from one another or from target objects over time.

\noindent\underline{T4: Multi-robot dynamic spatial reasoning.}
T4 level evaluates high-level collaborative reasoning over the full robot team. It tests whether the model can assess collective scene coverage, detect consistency or conflict among robot observations, identify the same object across views, reason counterfactually about alternative coordination strategies, update team-level beliefs from new observations, assign fetch tasks based on visibility, distance, and action context, and infer the overall team event, including acting and observing robots.
Overall, the taxonomy provides a structured evaluation of multi-robot egocentric reasoning, progressing from single-view perception to pairwise relations, global scene understanding, and dynamic team coordination.

\noindent\textbf{Baselines.}
\underline{Zero-shot MLLMs:}
We evaluate representative image/video MLLMs, including Qwen2.5-VL-7B/72B~\cite{bai2025qwen25vl}, Qwen3-VL-8B~\cite{bai2025qwen3vl}, InternVL3-8B~\cite{wang2025internvl3}, LLaVA-NeXT-Video-7B~\cite{zhang2024llavanextvideo}, and VideoLLaMA3-7B/2B~\cite{zhang2025videollama}, to assess their zero-shot performance on \textbf{CoopSR}.
\underline{Supervised fine-tuning baselines:}
We fine-tune the same major backbones with Low Rank Adaptation (LoRA)~\cite{hu2022lora} on the MLLM decoder on \textbf{CoopSR} to test whether standard task adaptation is sufficient and to isolate the architectural gains of \textbf{SG-CoR}.
\underline{Retrieval-augmented baselines:}
We compare with VideoRAG~\cite{luo2024video} and EgoRAG~\cite{egolife} to evaluate whether retrieved video or egocentric evidence improves reasoning over long, redundant multi-robot streams.
\underline{Keyframe-selection baselines:}
We include TSPO~\cite{tang2026tspo} and KTV~\cite{song2026ktv} under the best model from Supervised Fine-Tuning (SFT) to examine the performance of keyframe selection approaches.
Together, these baselines control for \emph{model capability}, \emph{task adaptation}, and \emph{evidence selection}, allowing us to verify that \textbf{SG-CoR} improves \textbf{CoopSR} through cooperative inductive biases for agent identity, co-visibility, temporal dynamics, and multi-robot fusion.

\noindent\textbf{Dataset splits and evaluation metric.} 
\underline{Data splits.} 
For the dataset with $114,227$ QAs collected from Habitat~\cite{puig2023habitat} and iGibson~\cite{li2021igibson}, we use $119/12/23$ unique scenes for train/val/test for Habitat, while the iGibson counterpart uses $8/2/4$ scenes for train/val/test. 
All $13$ scenes of the real-world set are used as the test set, which contributes to $2,326$ QAs in total.
\underline{Multiple-choice QA (MC4).}
For all 19 QA types, models produce a single letter (A/B/C/D) and are scored
Primary metrics are MC4 Accuracy (overall and per task-difficulty/team-size level), including per-difficulty-tier accuracy, per-source accuracy
(non-iGibson \textit{vs.} iGibson), and $N$-robot generalization (train on $N=[2,3]$, evaluate on $N=4$). 
AVG denotes the averaged QA performance on all samples without grouping.

\section{Methodology}
\label{sec:method}

\subsection{Problem Formulation}
\begin{figure}[t!]
\includegraphics[width=\linewidth]{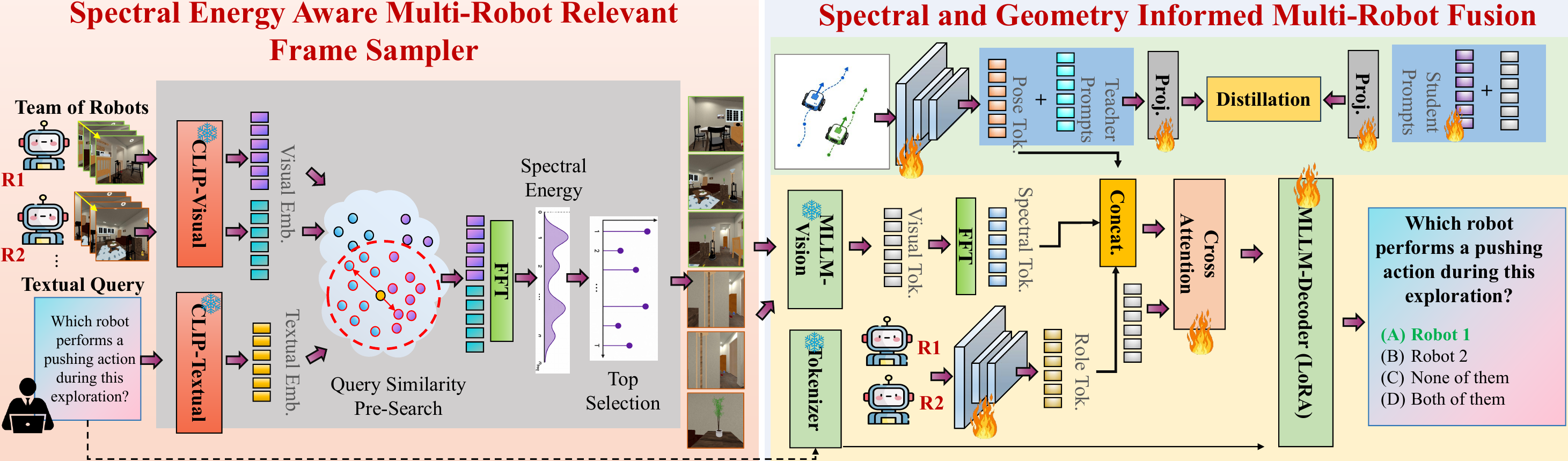}
\vskip-1ex
    \caption{Overview of \textbf{SG-CoR}, which combines query-guided spectral frame sampling, spectral- and geometry-informed multi-robot fusion, and prompt-space distillation for cooperative spatial reasoning on \textbf{CoopSR}.}
    \vskip-3ex
    \label{fig:main}
\end{figure}

We study \emph{multi-robot egocentric cooperative spatial reasoning}, where $N$ robots jointly answer a query from synchronized egocentric video streams. Robot $r$ observes $\mathcal{V}_r=\{\mathbf{x}_{r,t}\}_{t=1}^{T}$ and receives controls $\mathbf{u}_{r,t}$. During training, we additionally use pose estimates $\hat{\mathbf{u}}_{r,t}=(\hat{\mathbf{p}}_{r,t},\hat{\mathbf{e}}_{r,t})\in\mathrm{SE}(3)$ from simulators or motion capture; these poses are unavailable at test time.
Given $\mathcal{V}=\{\mathcal{V}_r\}_{r=1}^{N}$ and query $\mathbf{q}$, the model predicts
$\mathbf{y}=f_\theta(\mathcal{V},\{\hat{\mathbf{u}}_{r,t}\},\mathbf{q})$ during training,
$\mathbf{y}=f_\theta(\mathcal{V},\mathbf{q})$ during testing. $T$ and $N$ denote number of frames and robots. The active robot count $N$ and episode-local robot indices
are provided as input metadata. The robot index is passed to the model associated with its video stream.
This setting requires selecting sparse evidence across robots and time, resolving redundant or conflicting viewpoints, and reasoning over shared physical constraints.
\subsection{Spectral and Geometry-Aware Cooperative Reasoner (SG-CoR)}
Current MLLMs struggle with multi-robot cooperative spatial reasoning due to limited cross-robot dynamics modeling and weak use of physical priors. We propose \textbf{SG-CoR} (\textit{Spectral and Geometry-Aware Cooperative Reasoner}), which combines spectral energy-aware frame sampling with geometry-informed view fusion and prompt distillation. It selects informative multi-robot frames, injects motion and spatial priors during training, and enables pose-free cooperative reasoning at inference. The whole pipeline of our approach is shown in Figure~\ref{fig:main}.

\noindent\textbf{Spectral Energy-aware Multi-Robot Relevant Frame Sampler (SE-MR$^2$FS).}
Feeding all frames from all robots yields $\mathcal{O}(NT)$ visual tokens for $N$ robots and $T$ frames, which is costly and redundant due to adjacent near-duplicate frames and overlapping robot views. We independently compress each active robot stream to
$K \ll T$ frames using a training-free two-stage sampler: a frozen vision--language model first retrieves query-relevant candidates, which are then refined by temporal frequency energy over visual features.
Specifically, each frame is embedded with a frozen CLIP~\cite{radford2021learning} image encoder $\phi_v$ as $\mathbf{z}_{r,t}=\phi_v(\mathbf{x}_{r,t})\in\mathbb{R}^{d_c}$, while the query is encoded by the corresponding CLIP text encoder as $\mathbf{q}\in\mathbb{R}^{d_c}$. Both embeddings are $\ell_2$-normalized.

\noindent\underline{CLIP-based semantic candidate selection.}
For each robot $r$, we score every frame by its query similarity according to Eq.~\ref{eq:1}.
\begin{equation}
\label{eq:1}
    \mathbf{s}_{r,t} \;=\; \langle \mathbf{z}_{r,t},\, \mathbf{q}\rangle \;\in\; \left[-1, 1\right],
\end{equation}
and retain the top-$M$ frames as a candidate set as shown in Eq.~\ref{eq:2}.
\begin{equation}
\label{eq:2}
    \mathcal{C}_r \;=\; \mathrm{TopM}\!\big(\{\mathbf{s}_{r,t}\}_{t=1}^{T};\; M\big).
\end{equation}
This step removes frames that are obviously irrelevant to the question before any further step.

\noindent\underline{Temporal FFT energy on visual features.}
Many questions depend on understanding temporal dynamics, such as which robot moved first, when an object was placed, or which robot interacted with another object or agent. These changes cannot be captured reliably by static CLIP~\cite{radford2021learning} similarity alone. We therefore compute a per-frame motion descriptor by applying a discrete Fourier transform along the temporal axis of CLIP visual features. For a sliding window $\mathcal{W}(t)$ of length $w$ centered at $t$, let $\mathbf{Z}_{r,\mathcal{W}(t)} \in \mathbb{R}^{w \times d_c}$ denote the stacked CLIP features after subtraction of the local mean. Fourier transformation is conducted according to Eq.~\ref{eq:3}.
\begin{equation}
\label{eq:3}
    \Phi_{r,t}[k] \;=\; \log\!\Big(1 + \tfrac{1}{d_c}\big\lVert \mathcal{F}_w\!\big[\mathbf{Z}_{r,\mathcal{W}(t)}\big]_{k} \big\rVert_2^{2}\Big),\qquad k=\left[1,\dots,B\right],
\end{equation}
where $\mathcal{F}_w$ is the length-$w$ rFFT, the zero-frequency component is discarded so that the descriptor responds only to change, and the remaining bins are pooled into $B$ bands. 
The scalar frequency energy of frame $(r,t)$ is the sum of retained bands according to Eq.~\ref{eq:4}.
\begin{equation}
\label{eq:4}
    \mathbf{e}_{r,t} \;=\; \sum_{k=1}^{B} \Phi_{r,t}[k].
\end{equation} 
Because the FFT operates on CLIP embeddings rather than pixels, $\mathbf{e}_{r,t}$ captures semantically meaningful motion while remaining insensitive to nuisance pixel-level perturbation.
We restrict the FFT energy to the candidate set $\mathcal{C}_r$ and standardize it within that set to get $\tilde{\mathbf{e}}_{r,t}$. The final per-frame score combines frequency energy with an optional semantic-refinement term according to Eq.~\ref{eq:5},
\begin{equation}
\label{eq:5}
    \rho_{r,t} \;=\; \tilde{\mathbf{e}}_{r,t} \;+\; \tilde{\mathbf{s}}_{r,t},
    \qquad t \in \mathcal{C}_r,
\end{equation}
and the $K$ frames with highest $\rho_{r,t}$ within $\mathcal{C}_r$ are retained. 

\noindent\underline{Spectral token extraction.}
The same FFT processing yields a compact robot-level descriptor that is later injected into the fusion stage. After frame sampling, each robot supplies $K$ tunable visual tokens $\{\mathbf{h}_{r,k}\}_{k=1}^{K}$ produced by MLLM visual encoder. 
We compute the magnitude spectrum of this short sequence, retain its low-frequency coefficients, and project them through a small MLP with LayerNorm to obtain $\mathbf{F}_r \in \mathbb{R}^{d_F}$. $\mathbf{F}_r$ encodes how each robot's evidence evolves over the sampled subset, and is consumed by the fusion module as a per-robot temporal summary.

\noindent\textbf{Spectral and Geometry-Informed Multi-Robot Fusion (SGI-MRF).}
Single-robot video understanding mainly relies on temporal context within one coherent viewpoint. In contrast, multi-robot reasoning requires integrating geometrically constrained and mutually dependent observations. Robots may view the same object from different angles, suffer different occlusions, or share overlapping fields of view, so the model must reason about co-visibility, relative pose, and observation reliability.

Naive fusion methods such as concatenation, addition, or self-attention over pooled tokens treat robot observations as independent and interchangeable. 
They therefore cannot reliably resolve viewpoint conflicts, infer shared scene structure, or update team-level beliefs when observations disagree.
SGI-MRF is designed to explicitly model these cooperative dependencies.

Given the sampled visual tokens $\{\mathbf{h}_{r,k}\}_{k=1}^{K}$,
we construct a robot-level visual state $\mathbf{\mu}_r$ and a learned
uncertainty vector $\mathbf{\sigma}_r$ to support reliability-aware fusion. To capture temporal dynamics, we compute a Fourier embedding $\mathbf{F}_r$ from the sampled tokens, and encode robot motion commands into a pose embedding $\mathbf{P}_r$. We then form a geometric- and motion-aware robot token by combining pose, spectral, and robot-role embeddings as Eq.~\ref{eq:all}, using linear projection layers $\mathbf{W}_p$ and $\mathbf{W}_f$. Here, $\mathbf{E}_r$ is a learned robot-role embedding indexed by
the episode-local identity of Robot $r$. It provides an
explicit robot-identity cue and is added to the projected
motion and spectral embeddings to form the auxiliary token
$\mathbf{A}_r$. Robot indices are local to each episode and do not
identify a fixed physical robot across episodes.
\begin{equation}
\label{eq:all}
\mathbf{A}_r = \mathbf{W}_p \mathbf{P}_r + \mathbf{W}_f \mathbf{F}_r + \mathbf{E}_r.
\end{equation}
The visual states attend to these auxiliary tokens through cross-attention~\cite{chen2021crossvit} according to Eq.~\ref{eq:ca}.
\begin{equation}
\label{eq:ca}
\hat{\mathbf{\mu}}_{1:N}=\mathrm{CrossAttention}(\mathbf{\mu}_{1:N}, \mathbf{A}_{1:N}, \mathbf{A}_{1:N}),
\end{equation}
allowing each robot's view to be updated with motion, identity, and temporal cues. 

The uncertainty used for pooling is estimated from sampled
visual features, rather than simulator metadata or robot poses.
Let $\mathbf{h}_{r,k}$ denote the projected feature of the $k$-th sampled
frame from robot $r$. A learned temporal query produces
nonnegative attention weights $\mathbf{a}_{r,k}$ satisfying
$\sum_{k=1}^{K} \mathbf{a}_{r,k}=1$. We compute the attention-weighted
visual mean and temporal variance as
\begin{equation}
\label{eq:8}
\mathbf{\mu}_r^{\mathrm{vis}}
    = \sum_{k=1}^{K} \mathbf{a}_{r,k} \mathbf{h}_{r,k},  \mathbf{v}_r= \sum_{k=1}^{K} \mathbf{a}_{r,k}
       \left(\mathbf{h}_{r,k}-\mathbf{\mu}_r^{\mathrm{vis}}\right)^{2}.
\end{equation}
The initial visual uncertainty is
\begin{equation}
\mathbf{\sigma}_r^{\mathrm{vis}}
=
\operatorname{Softplus}
\left(\mathbf{g}_{\mathbf{\sigma}}\left(\mathbf{\mu}_r^{\mathrm{vis}}\right)\right)
+
\operatorname{sg}\left(\sqrt{\mathbf{v}_r}\right),
\end{equation}
where $\mathbf{g}_{\mathbf{\sigma}}$ is a learned uncertainty head,
$\operatorname{sg}$ denotes stop-gradient, and the square,
square root, and Softplus operations are elementwise.
The learned component models sample-dependent uncertainty,
while the empirical component measures temporal disagreement
among the selected frames.

SGI-MRF further projects this uncertainty and incorporates
a spectral contribution:
\begin{equation}
\bar{\mathbf{\sigma}}_r
    = \operatorname{Softplus}
       \left(\mathbf{W}_{\sigma}\sigma_r^{\mathrm{vis}}\right), \mathbf{\sigma}_r = \bar{\sigma}_r + 0.1\,\tilde{\mathbf{e}}_r\mathbf{1},
\end{equation}
where $\mathbf{W}_{\sigma}$ is a learned projection,
$\tilde{\mathbf{e}}_r$ denotes the normalized norm of the robot-level
Fourier embedding $F_r$, and $\mathbf{1}$ has the same
dimension as $\bar{\sigma}_r$.

The reliability weights underlying are
\begin{equation}
\mathbf{w}_r =
\frac{\exp\left(-\lVert\sigma_r\rVert_2\right)}
{\sum_{j=1}^{N}\exp\left(-\lVert\sigma_j\rVert_2\right)}.
\end{equation}

The implementation additionally applies norm clipping for
numerical stability. These weights constitute a learned
reliability-aware pooling mechanism, rather than a closed-form
Bayesian posterior. The fused visual states are aggregated
into the global team belief using these weights and the
projection $\mathbf{W}_b$, as shown in Eq.~\ref{eq:team_belief}.
\begin{equation}
    \mathbf{z}
    =
    \mathbf{W}_b
    \left(
        \sum_{r=1}^{N} \mathbf{w}_r \hat{\mu}_r
    \right).
    \label{eq:team_belief}
\end{equation}

\noindent\textbf{Geometry- and Motion-Informed Prompt-Space Distillation (GMIPSD).}
We propose GMIPSD, a privileged-information learning framework that uses simulator-provided ground-truth robot poses during training but requires no pose input at inference. We construct geometry- and motion-informed teacher prompts from robot poses to encode each robot's state and inter-robot spatial configuration. Learnable student prompts are then trained to align with these teacher prompts, distilling pose-induced physical awareness into task-adaptive latent representations. 
In inference, only the student prompts are used, enabling geometry- and motion-informed multi-robot fusion from egocentric videos alone.
The fused vector $\mathbf{z}$ conditions the language backbone $\mathrm{LM}_\phi$ via projected prefix tokens, which are physically aligned through distillation from a trajectory-aware teacher used only at training time. Both teacher and student output $L$ prompt vectors of dimension $d$, add learned positional codes, and apply LayerNorm. The teacher uses privileged inputs, \textit{i.e.}, ground-truth pose embeddings $\bar{\mathbf{p}}^{\mathrm{emb}}$ and learnable prefix tokens $\bar{\mathbf{C}}$ of the teacher, to emit a physically informed prompt as Eq.~\ref{eq:teacher}.
\begin{equation}
\label{eq:teacher}
    \mathbf{P}_T \;=\; \mathbf{g}_T\!\big(\bar{\mathbf{p}}^{\mathrm{emb}},\bar{\mathbf{C}}\big) \;\in\;\mathbb{R}^{L\times d}.
\end{equation}
The student counterpart has access only to test-time quantities, \textit{i.e.}, the global belief $\mathbf{z}$ and emits $\mathbf{P}_S = \mathbf{g}_S(\mathbf{z}, \mathbf{C})$, where $\mathbf{C}$ are learnable student prompts.
Here, $\mathbf{C}$ denotes learnable student prompt tokens, whereas
$\mathbf{C}_S \in \mathbb{R}^{L \times d}$ denotes the learned student
positional embeddings shown as grey blocks in Figure~\ref{fig:main}.
Writing the prompt generator before positional addition and
normalization as $g_S^{\mathrm{raw}}$, the student output is
\begin{equation}
    \mathbf{P}_S = \operatorname{LayerNorm}
\left(g_S^{\mathrm{raw}}(\mathbf{z},\mathbf{C}) + \mathbf{C}_S\right).
\end{equation}

Thus, $\mathbf{g}_S$ includes both positional addition and LayerNorm.
The projected student prompts $\mathbf{W}_P \mathbf{P}_S$ are concatenated with
the projected global team belief $\mathbf{W}_z \mathbf{z}$ to form the decoder prefix. Teacher prompts $\mathbf{P}_T$ provide the
distillation target and are not included in the decoder prefix.
The $\mathbf{P}_S$ and $\mathbf{P}_T$ are aligned with a distillation loss $L_{distill}$ formulated by cosine similarity distance.
In inference, the teacher is discarded; the student alone produces the prompt that conditions the language model, as shown in Eq.~\ref{eq:10}.
\begin{equation}
\label{eq:10}
    \tilde{\mathbf{H}} \;=\; \big[\,\mathbf{W}_z \mathbf{z};\; \mathbf{W}_P \mathbf{P}_S;\; \mathrm{Embed}(\mathbf{\hat{q}})\,\big],\qquad
    \mathbf{y} \;=\; \mathrm{LM}_\phi(\tilde{\mathbf{H}}),
\end{equation}
where $\mathbf{W}_z$ and $\mathbf{W}_P$ are linear projection layers. Embed indicates the tokenizer and $\hat{\mathbf{q}}$ denotes original query.

Because distillation happens in prompt space rather than output space, the student is supervised to recover the physical latent structure that would have best conditioned the LM, not merely its final answer -- this proves markedly more sample-efficient than standard response distillation.

\noindent\textbf{Training objective.}
We train end-to-end with a composite objective:
\begin{equation}
    \mathcal{L} \;=\; \lambda_{\mathrm{LM}}\mathcal{L}_{\text{LM}} \;+\; \lambda_d\,\mathcal{L}_{\text{distill}},
\end{equation}
where $\mathcal{L}_{\text{LM}}$ is the next-token cross-entropy on $\mathbf{y}$ and $\mathcal{L}_{\text{distill}}$ is the geometry- and motion-informed prompt distillation loss. The loss weights $\lambda_{LM}$ and $\lambda_{d}$ are selected using grid search.

\section{Experimental Results}

\noindent\textbf{Implementation details.} 
All experiments are conducted on $4$ NVIDIA A100 GPUs. We precompute CLIP and Qwen visual features before training. The model is trained for $10$ epochs using AdamW~\cite{loshchilov2017decoupled} with learning rate $1 \times 10^{-4}$, per-GPU batch size $16$, gradient accumulation $4$, and gradient clipping $1.0$. 
Learnable prompts number is $8$, $M$, $w$ and $K$ are selected as $32$, $4$ and $8$, LoRA fine-tuning on MLLM decoder is applied with rank $8$, $\alpha=16$, and dropout $0.05$. $\lambda_{LM}$ and $\lambda_{d}$ are selected as $1.0$ and $0.3$. Hyperparameters are selected through grid searching and validation set performance. The number of trainable parameters of \textbf{SG-CoR} with Qwen2.5-VL-7B~\cite{hui2024qwen2} is 23.8M.

\label{subsec:coopsr_results}
\begin{table}[t!]
\caption{CoopSR benchmark results.
We report accuracy for four reasoning tiers (T1--T4), different team sizes ($N=2,3,4$), and the overall average. Note: AVG is computed over all QA samples, rather than by averaging across task types or robot-team sizes. Team-size results compare independently generated episode sets
with matched collection protocols; they do not constitute a
trajectory-controlled comparison of adding robots.}
\label{tab:coopsr}
\centering
\resizebox{\columnwidth}{!}{
\begin{tabular}{l|llll|lll|l|llll|lll|l}
\toprule
\textbf{Datasets}  & \multicolumn{8}{c|}{\textbf{Habitat}~\cite{puig2023habitat}}                                                               & \multicolumn{8}{c}{\textbf{iGibson}~\cite{li2021igibson}}                                  \\
\textbf{Models}                 & \textbf{T1} & \textbf{T2} & \textbf{T3} & \textbf{T4} & \textbf{N=2} & \textbf{N=3} & \textbf{N=4} &\textbf{AVG} & \textbf{T1} & \textbf{T2} & \textbf{T3} & \textbf{T4} & \textbf{N=2} & \textbf{N=3} & \textbf{N=4} & \textbf{AVG} \\
\midrule
\multicolumn{17}{c}{\textit{\textbf{Zero-shot evaluations}}}\\                    
\midrule                                                         
Qwen2.5-VL-7B-Instruct~\cite{bai2025qwen25vl} & 23.57&39.61  &28.40  &34.35  &34.66 & 34.99    & 31.03 &    33.51    &  39.44   &  50.89     &  28.12     &   29.21 & 31.38  &35.61 &34.84 & 33.95   \\
Qwen2.5-VL-72B-Instruct~\cite{bai2025qwen25vl}& 18.84& 43.57 & 19.82 &33.61  & 30.91 &  33.18  & 31.15 & 31.75&18.84  & 43.57 & 19.82  & 33.61&27.45&30.18 &33.18 &30.93\\
LLaVA-NeXT-Video-7B~\cite{zhang2024llavanextvideo} &26.76 &36.35  &33.30  &27.09  &31.82 &31.49     &30.57        &  31.27 &  21.11   &  30.95     &  32.37     &   31.32    &  27.90 &33.08  &31.71 & 31.27   \\
InternVL3-8B~\cite{wang2025internvl3} &20.77 & 49.86 & 36.08 & 43.01 & 41.03 & 42.04    &  41.57 &  41.56 & 28.15 &  52.25 &  34.80 &  36.34 &  42.07    & 38.61& 38.23 & 38.84 \\
VideoLLaMA3-7B~\cite{zhang2025videollama} & 23.96 & 44.59 & 37.86 & 44.03 & 40.84 & 42.16    & 40.52  &  41.17  & 19.44 &  46.28 &  39.21 &  40.09  & 37.51  &40.30 & 42.22&  40.05\\
VideoLLaMA3-2B~\cite{zhang2025videollama} & 21.26& 41.69 & 35.10 & 37.83 & 33.77& 38.02    &  39.08  &  37.04 & 12.78 &  55.51  &  31.61 &  42.64  & 40.90  & 40.67& 42.66 & 41.44 \\
Qwen3-VL-8B-Instruct~\cite{bai2025qwen3vl}  &20.77 & 45.65 & 37.32 &39.25  &42.18 & 39.35  & 36.46  & 39.24 & 35.56 &  47.72  & 37.66 &  35.54& 37.60  & 40.21 & 35.33&  38.46  \\
\midrule
\multicolumn{17}{c}{\textit{\textbf{SFT baselines}}}       \\                   
\midrule                                                                                       
Qwen2.5-VL-7B~\cite{bai2025qwen25vl} & 46.67 & 76.30 & 53.02 & 72.53 & 75.04& 63.82    &  61.77  &  66.68 & 80.00 &   65.77 &  54.10 &   64.74 & 63.40   & 63.92& 63.68& 63.67   \\
Qwen2.5-VL-72B~\cite{bai2025qwen25vl} & 43.96& 74.47 & 49.90 & 68.66 & 71.05&  60.88   &  59.56  & 63.65 & 73.89 &70.39 & 49.09   & 60.24 & 58.19 &63.17 & 61.16 & 60.83 \\
LLaVA-NeXT-Video-7B~\cite{zhang2024llavanextvideo} &  20.97& 55.43 & 47.35 & 49.25 & 56.19 &50.91 &48.42  &  51.72  &  18.33  &  53.27 &  47.57  & 49.25      &  45.01 & 49.11& 49.96&  48.05\\
InternVL3-8B~\cite{wang2025internvl3}  & 39.03 & 75.51 & 47.60 & 68.97 & 69.86 & 61.76   & 58.25  &   63.12 & 79.44 &  69.35 &   47.60    &  68.97  & 57.09& 62.61 &58.56 & 59.38\\
VideoLLaMA3-7B~\cite{zhang2025videollama} & 19.52&  62.58& 42.91 & 63.22 & 61.66 & 53.06    &  49.85& 54.68 &  13.89 &  55.95 &  47.72 &  51.25 &  43.55    & 43.61& 51.26& 49.47\\
VideoLLaMA3-2B~\cite{zhang2025videollama} & 20.68 & 59.42 & 42.63 & 63.60 & 61.09&  52.40   &   48.92 &  53.95& 15.56 &  53.12 &  44.68 &   51.92 & 45.11 &50.05 & 51.00& 48.75  \\
Qwen3-VL-8B-Instruct~\cite{bai2025qwen3vl} & 30.53& 67.83 & 46.83 & 67.55 & 65.95 & 57.90    &  55.49  &  59.62 & 59.44 &  57.44     &   48.33    & 54.69  & 51.33 & 53.80 & 57.43 & 54.24 \\
\midrule
\multicolumn{17}{c}{\textit{\textbf{RAG Baselines (Best SFT model adopted)}}}       \\
\midrule
VideoRAG~\cite{jeong2025videorag} (Qwen2.5-VL-7B-Instruct~\cite{bai2025qwen25vl})  & 39.13& 76.52 & 51.08 & 71.37 &71.34 & 63.19    & 61.68  &  65.25 &  67.78 &   70.68    &  51.82  &  57.86 & 58.01 & 60.26 & 61.08 & 59.80   \\
VideoRAG~\cite{jeong2025videorag} (InternVL3-8B~\cite{wang2025internvl3})   &22.03 & 72.92 & 45.46 & 67.00&  66.69 & 57.92 & 55.53 &59.88 &12.22 & 64.14 & 45.90&  50.53 & 47.94 & 52.39 & 50.56 &50.29\\
EgoRAG~\cite{egolife} (Qwen2.5-VL-7B-Instruct~\cite{bai2025qwen25vl}) & 48.31 & 74.93 & 51.80 &  71.36& 73.51&  63.61   & 60.37       &  65.63  &  77.78   &   68.15    &  54.41  &  64.02  &  61.94  & 64.14 & 64.03&  63.70  \\
EgoRAG~\cite{egolife} (InternVL3-8B~\cite{wang2025internvl3})  &  39.90 &  74.71 & 47.40   &68.47   &  69.77  &    60.67    &     58.20   &  62.70      &   72.22     &  69.05  & 44.98       &   58.41     &  56.91  &  60.82   & 58.30 & 58.65  \\
\midrule
\multicolumn{17}{c}{\textit{\textbf{Keyframe Baselines (Best SFT model adopted)}}}       \\
\midrule
TSPO~\cite{tang2026tspo} (Qwen2.5-VL-7B-Instruct~\cite{bai2025qwen25vl})   & 37.97  &  75.80  &  46.68  &  68.55  & 69.28  &  61.44  &  57.94  &  62.72  &  \textbf{86.11} &   67.56  &  45.29  &  56.47  &  55.54  &  60.45  &  58.38 &   58.11  \\
TSPO~\cite{tang2026tspo} (InternVL3-8B~\cite{wang2025internvl3})  &  42.51  &76.21 & 50.77 & 70.79 & 72.21 & 63.25 & 60.35 & 65.09 & 79.44 &  72.02 & 49.54 &  63.13 &  62.21 &  64.67 & 62.55 & 63.12 \\
KTV~\cite{song2026ktv} (Qwen2.5-VL-7B-Instruct~\cite{bai2025qwen25vl})  &  19.71   & 52.32   &  41.68    &  57.11  &  50.56    &   49.63   &    47.33 & 49.13  &  21.67     &  53.12    &   43.47     &  42.03     &  38.98     & 45.55 & 45.79  & 43.46\\
KTV~\cite{song2026ktv} (InternVL3-8B~\cite{wang2025internvl3})  &  19.13   & 61.43 & 40.26 & 62.95 &60.60   & 51.92&  48.61  &53.53  &  13.89     &   58.63   &   40.88     &   43.25    &  42.27     & 46.58  & 44.14 & 44.31\\
\midrule
\multicolumn{17}{c}{\textit{\textbf{Our proposed approach}}}       \\
\midrule
\textbf{SG-CoR} (InternVL3-8B~\cite{wang2025internvl3})& \textbf{57.78} & 78.31 & 56.98 &  \textbf{74.96} & 77.07 & \textbf{68.07} & 65.37 & 69.99 & 76.11 & \textbf{73.81} & 55.17 & 73.51& 72.28 & 68.32 & 69.59 &70.07 \\
\textbf{SG-CoR} (Qwen3-VL-8B-Instruct~\cite{bai2025qwen3vl})
& 57.20&   78.14 &  58.38 &  74.54  &  \textbf{78.28} &  67.32 &  65.31&  70.11   & 79.44  & 72.77 & \textbf{59.42} &  73.18 &  72.37 & \textbf{70.29} & 69.50 & 70.70\\
\textbf{SG-CoR} (Qwen2.5-VL-7B-Instruct~\cite{bai2025qwen25vl}) & 55.36 & \textbf{78.72} & \textbf{59.36} & 74.94 & 78.26 & 67.34 & \textbf{66.60} & \textbf{70.55} & 74.44 & 71.88 & 58.36  & \textbf{74.63} & \textbf{72.74} & 69.92 & \textbf{69.85} & \textbf{70.82} \\
\bottomrule
\end{tabular}
}
\vskip-2ex
\end{table}

\begin{figure}[t!]
\includegraphics[width=\linewidth]{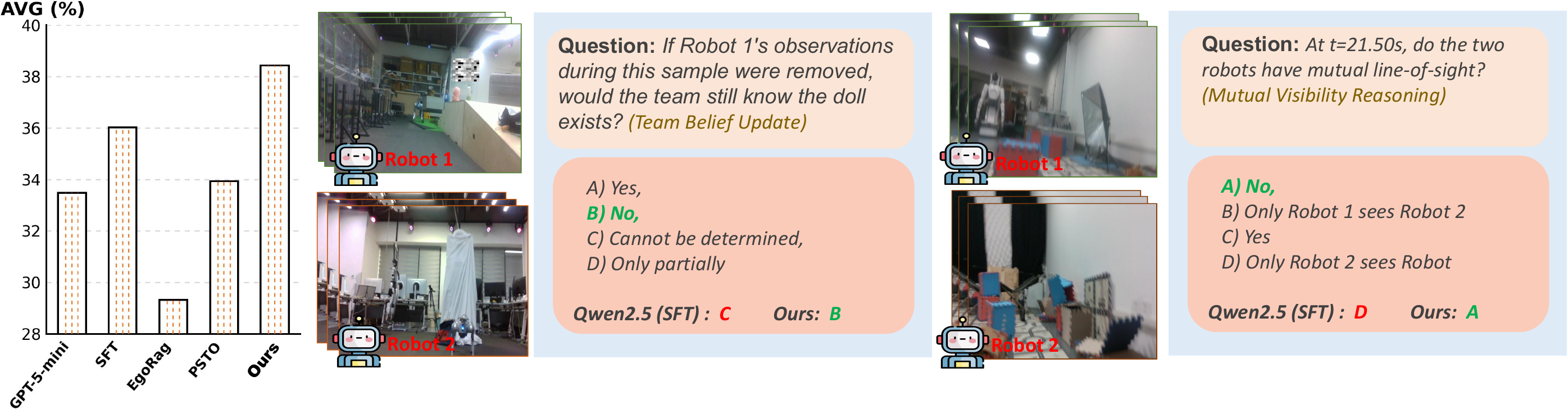}
\vskip-1ex
\caption{Performances of MLLMs on the real-world test set. Histograms on the left show the accuracy of different MLLMs. Samples on the right are qualitative samples.}
\vskip-3ex
\label{fig:real}
\end{figure}
\noindent\textbf{Analysis of the benchmark.}
Table~\ref{tab:coopsr} compares different model groups on \textbf{CoopSR} across Habitat~\cite{puig2023habitat} and iGibson~\cite{li2021igibson}, covering four reasoning tiers (T1--T4) and team sizes $N\in\{2,3,4\}$.
Zero-shot MLLMs obtain low average accuracies, mostly between $31\%$ and $42\%$, showing that general-purpose MLLMs struggle with cooperative spatial reasoning from multi-robot egocentric videos. 
SFT substantially improves performance, for example, Qwen2.5-VL-7B-Instruct~\cite{bai2025qwen25vl} increases from $33.51\%$ to $66.68\%$ on Habitat and from $33.95\%$ to $63.67\%$ on iGibson~\cite{li2021igibson}.
Accuracy differences across team sizes reflect variation in
both team configuration and generated episodes, and therefore
cannot be attributed solely to the number of robots.
\begin{table}[t!]
\caption{Ablation results on the Qwen2.5 backbone~\cite{bai2025qwen25vl} for Habitat and iGibson test sets.}
\label{tab:abl}
\centering
\resizebox{0.8\columnwidth}{!}{\begin{tabular}{lllllllll}
\toprule
\multirow{2}{*}{\textbf{Method}}                  & \multicolumn{4}{c}{\textbf{Habitat}~\cite{puig2023habitat}}   & \multicolumn{4}{c}{\textbf{iGibson}~\cite{li2021igibson}}   \\
                                         & N=2   & N=3   & N=4   & AVG   & N=2   & N=3   & N=4   & AVG   \\
\midrule
w/o Spectral Energy Ranking in SE-MR$^2$FS  & 77.63 & 65.52 & 64.55 &69.03 &67.80 & 69.73 & 67.85& 68.44 \\
w/o Fourier Embed. in SGI-MRF
& 73.95 & 64.65 & 62.57 & 66.88 & 59.93 & 61.39 & 60.03 & 60.43 \\
\midrule
w/o SE-MR$^2$FS  & 72.53 & 63.75 & 61.00 & 65.58 & 61.94 & 65.98 & 59.51 & 62.40 \\
w/o SGI-MRF &  75.23     &  64.87     &   62.22    &  67.24     &  51.42     &   59.04    &   55.00    &  55.12     \\
w/o GMIPSD &    73.89   &  63.69     &   61.70    &    66.24   &  60.57     &   63.26    &   61.60    &    61.79   \\
\midrule
Ours & \textbf{78.26} & \textbf{67.34} & \textbf{66.60} & \textbf{70.55} & \textbf{72.74} & \textbf{69.92} & \textbf{69.85} & \textbf{70.82} \\
\bottomrule
\end{tabular}}
\vskip-4ex
\end{table}
\begin{table}[t!]
\caption{Cross-team-size generalizability analysis on Habitat and iGibson when we train the model on $N=2$ and $N=3$ while we test on $N=4$. 
Note that the averaged ACC is calculated on the averaged performance of all QAs.}
\label{tab:teamsize}
\centering
\resizebox{0.8\columnwidth}{!}{\begin{tabular}{llllll|lllll}
\toprule
\multirow{2}{*}{\textbf{Method}}                  & \multicolumn{5}{c}{\textbf{Habitat}~\cite{puig2023habitat}}   & \multicolumn{5}{c}{\textbf{iGibson}~\cite{li2021igibson}}   \\
                                         & \textbf{T=1}   & \textbf{T=2}   & \textbf{T=3} &\textbf{T=4}  & \textbf{AVG}   & \textbf{T=1}   & \textbf{T=2}   & \textbf{T=3} & \textbf{T=4}   & \textbf{AVG}   \\
\midrule
SFT (Qwen2.5-VL-7B-Instruct~\cite{bai2025qwen25vl}) & 18.84 & 62.39 & 41.81 & 58.24& 52.30& 15.00& 63.44& 37.08& 48.24& 47.18\\
SFT (Qwen3-VL-8B-Instruct~\cite{bai2025qwen3vl}) &22.03 &60.65  &41.30  &58.68 &52.09 &41.67 &57.71 &45.42 & 53.85& 52.22\\
SFT (InternVL3-8B~\cite{wang2025internvl3}) & 29.86 & 69.49 & 47.32 & 62.81 & 58.23 & 45.00 & 66.96 & 42.92 &56.09 &54.91\\
\midrule
EgoRAG~\cite{egolife} (Qwen2.5-VL-7B-Instruct~\cite{bai2025qwen25vl}) &24.06 & 48.41 & 37.54 & 54.12 &46.12 &53.33 & 54.19 & 39.17&  45.35& 46.22 \\
EgoRAG~\cite{egolife} (Qwen3-VL-8B-Instruct~\cite{bai2025qwen3vl}) &22.32 & 47.17 & 39.86 & 53.97 &46.24& 21.67 & 38.77 & 47.50 & 39.90& 40.31 \\
EgoRAG~\cite{egolife} (InternVL3-8B~\cite{wang2025internvl3}) &19.13 & 53.91 & 39.35 & 56.38 &48.65 &26.67 & 59.47 & 35.42 & 43.91 &44.31 \\
\midrule
TSPO~\cite{tang2026tspo} (Qwen2.5-VL-7B-Instruct~\cite{bai2025qwen25vl}) & 23.33 & 63.19 & 38.19 & 62.83 &53.63  & 23.33 & 45.00 & 45.00&  58.64 &51.35 \\
TSPO~\cite{tang2026tspo} (Qwen3-VL-8B-Instruct~\cite{bai2025qwen3vl}) & 20.87  &51.81 & 40.58 & 56.82 &48.71 & 11.67 & 57.27 & 44.17 & 43.75 &44.83\\
TSPO~\cite{tang2026tspo} (InternVL3-8B~\cite{wang2025internvl3}) &20.58 & 56.96 & 39.71&  56.72 &49.80 &18.33  &59.47  &43.33  &52.24 &50.04 \\
\midrule
\textbf{SG-CoR} (Qwen2.5-VL-7B-Instruct~\cite{bai2025qwen25vl}) & 50.14 & 70.94 & 51.81 & 68.79& 63.56 & 71.67 & 66.96 & 51.25& 71.15& 66.20\\
\textbf{SG-CoR} (Qwen3-VL-8B-Instruct~\cite{bai2025qwen3vl}) &48.12 & 71.59 & \textbf{54.57} &  68.20& 64.11& 66.67 & 68.72 & \textbf{56.67} & 70.99 &67.33\\
\textbf{SG-CoR} (InternVL3-8B~\cite{wang2025internvl3}) & \textbf{51.59} & \textbf{72.39} & 53.48 & \textbf{69.53} & \textbf{64.79} & \textbf{81.67} & \textbf{70.48} & 55.42 & \textbf{73.08} & \textbf{69.33}\\
\bottomrule
\end{tabular}}
\vskip-3ex
\end{table}
RAG and keyframe baselines such as EgoRAG~\cite{egolife} and TSPO~\cite{tang2026tspo} further improve evidence selection, but they mainly operate at the input level and do not explicitly model robot identity, inter-robot geometry, or cooperative view fusion. 
In contrast, \textbf{SG-CoR} achieves the best overall performance across both environments. 
The strongest \textbf{SG-CoR} variant reaches $70.55\%$ on Habitat and $70.82\%$ on iGibson, outperforming the strongest non-SG-CoR baselines by $3.87\%$ and $7.12\%$, respectively.

These gains indicate that the key challenge in \textbf{CoopSR} is not only selecting relevant frames, but also forming a consistent multi-robot scene representation. \textbf{SG-CoR} addresses this with three complementary biases: spectral sampling captures informative temporal changes for action and visibility reasoning; geometry- and motion-informed fusion aligns observations across changing viewpoints; and prompt distillation transfers trajectory-level geometric supervision to the student model without requiring ground-truth poses at test time. Together, these components improve cross-view consistency and team-level reasoning, leading to stronger gains on T4.

\noindent\textbf{Analysis of ablation study.}
Table~\ref{tab:abl} ablates each component of \textbf{SG-CoR} with the Qwen2.5-VL-7B-Instruct backbone~\cite{bai2025qwen25vl}. 
The full model achieves the best average accuracy on both Habitat~\cite{puig2023habitat} and iGibson~\cite{li2021igibson}, with $70.55\%$ and $70.82\%$, respectively.
Removing any component consistently degrades performance, showing that spectral sampling, Fourier temporal encoding, spectral-aware fusion, and geometry- and motion-informed prompt distillation are all beneficial. 
The largest drop occurs on iGibson~\cite{li2021igibson} when SGI-MRF is removed, reducing accuracy from $70.82\%$ to $55.12\%$, highlighting the importance of robust cross-view fusion in complex interactive scenes. 
Overall, the ablation confirms that these components are complementary and jointly improve multi-robot cooperative spatial reasoning.

\noindent\textbf{Cross-team-size generalization.}
Table~\ref{tab:teamsize} evaluates generalization from smaller teams ($N=\left[2,3\right]$) to an unseen larger team size ($N=4$). This setting is challenging due to increased observation redundancy, viewpoint inconsistency, and more complex inter-agent relations. Across Habitat~\cite{puig2023habitat} and iGibson~\cite{li2021igibson}, \textbf{SG-CoR} achieves the best performance for all three backbones. The strongest \textbf{SG-CoR} variant reaches $64.79\%$ on Habitat, improving over the best baseline by $6.56\%$, and $69.33\%$ on iGibson~\cite{li2021igibson}, with a larger gain of $14.42\%$.
These results show that \textbf{SG-CoR} generalizes more effectively than SFT, EgoRAG~\cite{egolife}, and TSPO~\cite{tang2026tspo}. 
Beyond task adaptation, retrieval, or keyframe selection, \textbf{SG-CoR} combines dynamic spectral sampling with geometry- and motion-informed fusion and prompt distillation, explicitly modeling robot relations, co-visibility, and cross-view alignment. These cooperative inductive biases enable stronger transfer to the unseen $N=4$ setting.

\noindent\textbf{Real-world generalizability.}
Figure~\ref{fig:real} shows that \textbf{SG-CoR} achieves the best real-world accuracy, outperforming GPT-5 mini~\cite{singh2025openai}, SFT, EgoRAG~\cite{egolife}, and TSPO~\cite{tang2026tspo}. Qualitative examples further show that \textbf{SG-CoR} handles team belief updates and mutual visibility reasoning more reliably than SFT, indicating stronger generalizability to viewpoint changes, partial observations, and multi-robot spatial relations in real-world situations.

\section{Conclusion}
We introduced \textbf{CoopSR}, a benchmark for multi-robot cooperative egocentric spatial reasoning, requiring models to integrate synchronized views for spatial, temporal, visibility, and coordination reasoning. We further proposed \textbf{SG-CoR}, which combines spectral frame sampling, geometry- and motion-informed fusion, and prompt distillation to build coherent team-level representations without requiring privileged pose information at test time. 
Experiments on Habitat, iGibson, cross-team-size evaluation, and real-world robot data show that \textbf{SG-CoR} consistently outperforms strong MLLM, SFT, RAG, and keyframe baselines, highlighting the importance of physically grounded cooperative reasoning for embodied AI.

\clearpage
\bibliographystyle{plain}
\bibliography{main.bib}


\appendix

\section*{Acknowledgments and Disclosure of Funding
}
The project is funded by the Deutsche Forschungsgemeinschaft (DFG, German Research Foundation) – SFB 1574 – 471687386. This work was performed on the HoreKa supercomputer funded by the Ministry of Science, Research and the Arts Baden-Württemberg and by the Federal Ministry of Education and Research. The authors also acknowledge support by the state of Baden-Württemberg through bwHPC and the German Research Foundation (DFG) through grant INST 35/1597-1 FUGG. 
This project is also supported in part by the National Natural Science Foundation of China under Grant No. 62473139, No. 62673195, and No. 62388101, in part by the Hunan Provincial Research and Development Project (Grant No. 2025QK3019), and in part by the State Key Laboratory of Autonomous Intelligent Unmanned Systems (the opening project number ZZKF2025-2-10).
This research was partially funded by the Ministry of Education and Science of Bulgaria (support for INSAIT, part of the Bulgarian National Roadmap for Research Infrastructure).
\section{Appendices}
\subsection{Societal Impact and Limitations}
This work advances cooperative spatial reasoning for multi-robot teams that jointly interpret synchronized egocentric observations. By enabling robots to integrate partial viewpoints into a shared scene understanding, \textbf{CoopSR} may support important applications in search and rescue, assistive robotics, smart homes, warehouse automation, infrastructure inspection, and navigation in hazardous or unknown environments. In these settings, stronger team-level situational awareness can help robots cover larger areas, resolve occlusions, coordinate tasks more effectively, and reduce human exposure to risk.

At the same time, multi-robot egocentric perception introduces privacy and safety challenges. Robots operating in homes, workplaces, hospitals, or public spaces may capture sensitive visual information, so deployment should follow data minimization, consent, anonymization, secure storage, and access-control practices. In addition, higher benchmark accuracy should not be interpreted as a guarantee of safe autonomous behavior. False spatial reasoning, incorrect visibility estimation, or wrong task assignment could lead to unsafe navigation, collisions, failed coordination, or inappropriate actions in physical environments. Therefore, real-world deployment should include human oversight, uncertainty estimation, fail-safe mechanisms, and extensive validation under diverse operating conditions.

Our dataset is primarily simulation-based, with a controlled real-world quadruped robot test set. While this enables scalable and systematic evaluation, it may not fully capture open-world variation such as clutter, lighting changes, sensor noise, social context, larger robot teams, or unseen embodiments. Cooperative robot reasoning also has dual-use potential: it can improve rescue, accessibility, and industrial safety, but could also be misused for surveillance or intrusive monitoring. Future work should therefore evaluate robustness, privacy preservation, fairness, and human-centered safety alongside accuracy.

\paragraph{Limitation Discussion.}
Although \textbf{CoopSR} covers diverse simulated environments,
its real-world evaluation remains limited in scene diversity,
robot types, and team sizes. The benchmark deliberately controls
embodiment to isolate cooperative spatial reasoning, while the
Go2/Go2-W team introduces only limited within-family platform
variation. \textbf{SG-CoR} processes per-agent RGB streams without
requiring robot poses at inference, but this architectural
flexibility does not establish robustness to arbitrary
embodiments or sensing suites.

Future work should evaluate larger heterogeneous teams,
including aerial--ground teams and unseen embodiments, while
systematically varying camera intrinsics, field of view, sensor
quality, robot height, and locomotion type. Extending beyond RGB
observations would additionally require modality-specific
encoders. Broader evaluation should also encompass open-ended
tasks and safety-critical scenarios with richer physical
interactions.

\subsection{Details of Simulator and Dataset Splits}
\begin{figure}[h]
\includegraphics[width=\linewidth]{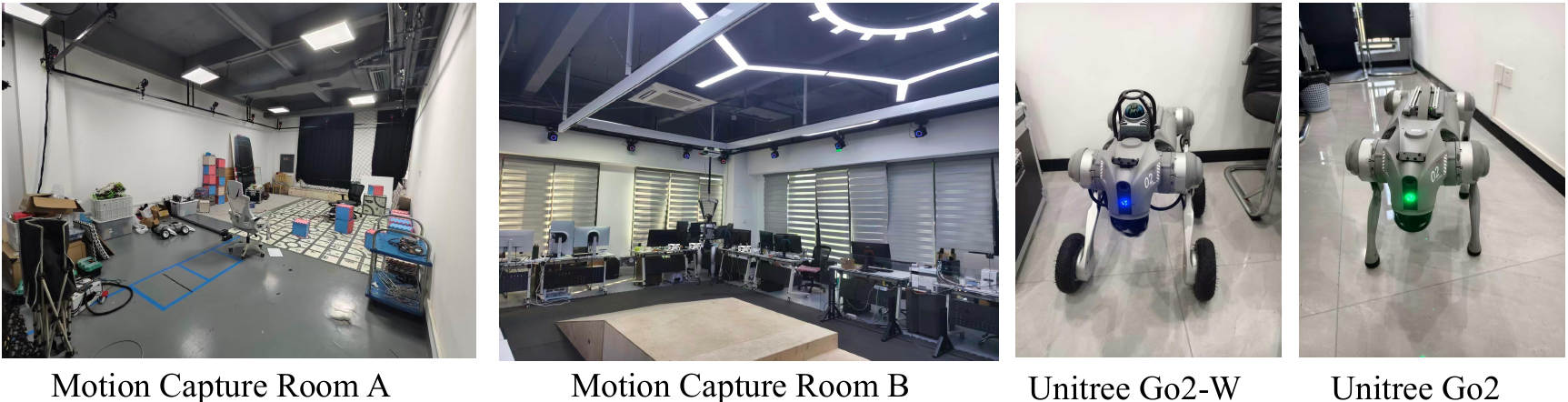}
    \caption{Overview of data collection devices. We make use of two Unitree quadruped robots and two different motion capture systems from FZMotion and NOKOV. 
    We use daily life objects to set up $13$ different scenarios in these two data collection rooms, which can provide ground truth pose data for the robots.}
    \label{fig:devices}
\end{figure}

\noindent\textbf{Habitat scenes.} 
Habitat~\cite{puig2023habitat} is a high-performance embodied AI simulation platform that supports photorealistic indoor navigation, egocentric RGB-D observation, and scalable agent-environment interaction.
Robots are deployed in $154$ procedurally generated indoor floor plans.
Each scene is explored by $N\in\{2,3,4\}$ robots executing a coverage exploration policy.
Synchronised RGB video, semantic segmentation mask for each frame, depth video, odometry commands, and ground-truth poses are recorded.
The trajectory log for each exploration is stored together with the co-located video files,
containing per-timestep columns: position, heading,
forward velocity, and angular velocity.

\noindent\textbf{iGibson scenes.} IGibson~\cite{li2021igibson} is a realistic interactive simulation platform for embodied AI, supporting indoor navigation, geometry- and motion-informed interaction, and egocentric RGB-D perception in diverse household environments. $14$ interactive scenes are adopted for the data collection. During the exploration, we also adopt $N \in \{2,3,4\}$ for the data collection.

\noindent\textbf{Train / val / test splits.}
For each configuration, scenes are divided into disjoint training, validation, and test splits, with $119/12/23$ scenes for Habitat settings and $8/2/4$ scenes for iGibson. 
For the real-world quadruped robot set, all collected scenes are used for testing. No scene is shared across splits, thereby preventing scene-level data leakage.

\noindent\textbf{Episode generation across team sizes.}
The scene name and exploration index initialize each simulator
run, after which robot starting positions, waypoints, and motions
are generated jointly for the specified team size. Habitat
computes rendezvous and tether targets from the mean position
of all active robots, while iGibson computes NEAR-phase waypoints
from the full-team centroid. Consequently, changing $N$ changes
the trajectories and observations even for the same scene and
exploration index. A future paired observation evaluation could
hold an $N=4$ episode fixed and mask predefined robot streams
to evaluate $N=\{2,3\}$, isolating the effect of available views
while preserving the underlying trajectories.

\subsection{Dataset Annotation Details, Diversity and Statistical Analysis}
\subsubsection{Human Checking Details}
To ensure the reliability of \textbf{CoopSR}, we conduct a multi-stage human verification process. Four human annotators independently review the generated QA pairs using synchronized multi-robot egocentric videos, simulator metadata, and the proposed answer choices. Annotators verify whether each question is visually grounded, whether the answer is uniquely determined, and whether the question genuinely requires cooperative reasoning across multiple robot viewpoints rather than single-view recognition or language priors.

For each QA pair, annotators either accept it, correct it, or discard it. Corrections include resolving robot identity confusion, inaccurate spatial relations, ambiguous object references, incorrect visibility states, and temporal boundary inconsistencies. QA pairs are discarded if the target object is not sufficiently observable, if multiple answer choices are plausible, if the answer can be inferred from textual priors alone, or if a cross-robot level question does not require cross-robot evidence despite belonging to a cooperative reasoning category.
To further improve benchmark quality, QA pairs marked ambiguous by at least two annotators are removed. 

\subsubsection{Dataset Diversity and Statistical Analysis}
\begin{figure*}[t!]
    \centering

    \begin{subfigure}{0.32\textwidth}
        \centering
        \includegraphics[width=\linewidth]{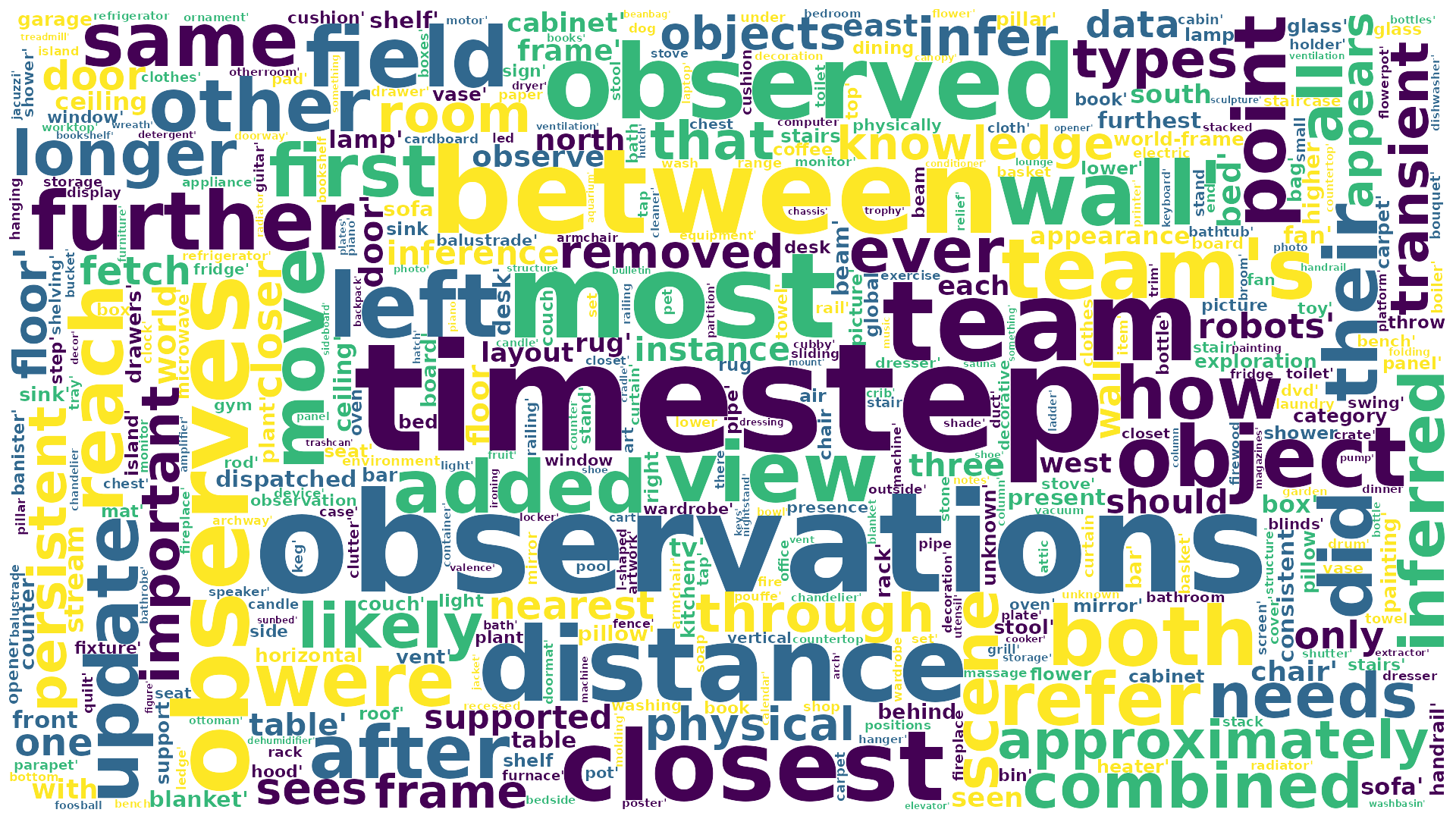}
        \caption{Habitat set~\cite{puig2023habitat}}
        \label{fig:image1_habitat}
    \end{subfigure}
    \hfill
    \begin{subfigure}{0.32\textwidth}
        \centering
        \includegraphics[width=\linewidth]{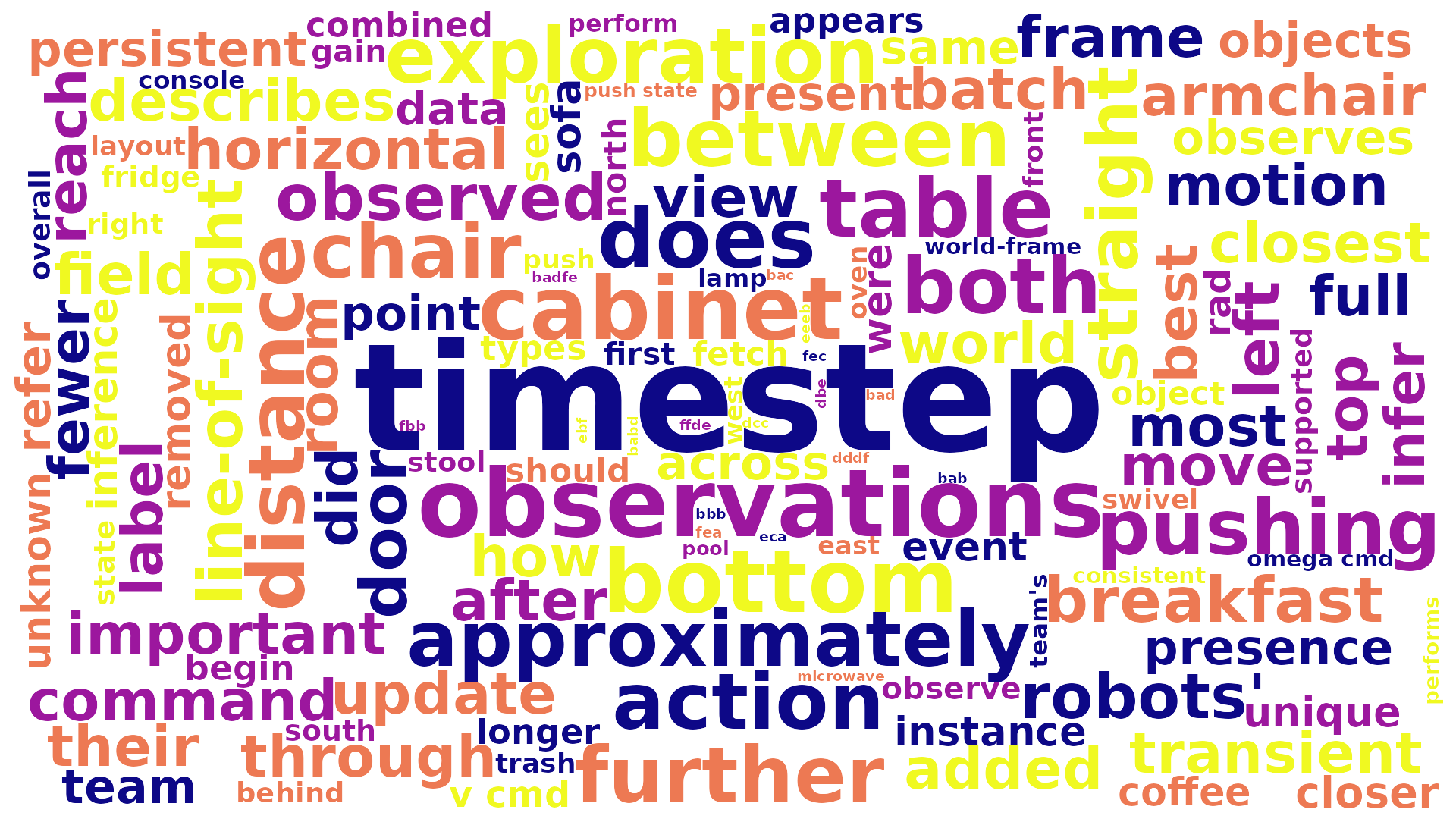}
        \caption{iGibson set~\cite{li2021igibson}}
        \label{fig:image2_igibson}
    \end{subfigure}
    \hfill
    \begin{subfigure}{0.32\textwidth}
        \centering
        \includegraphics[width=\linewidth]{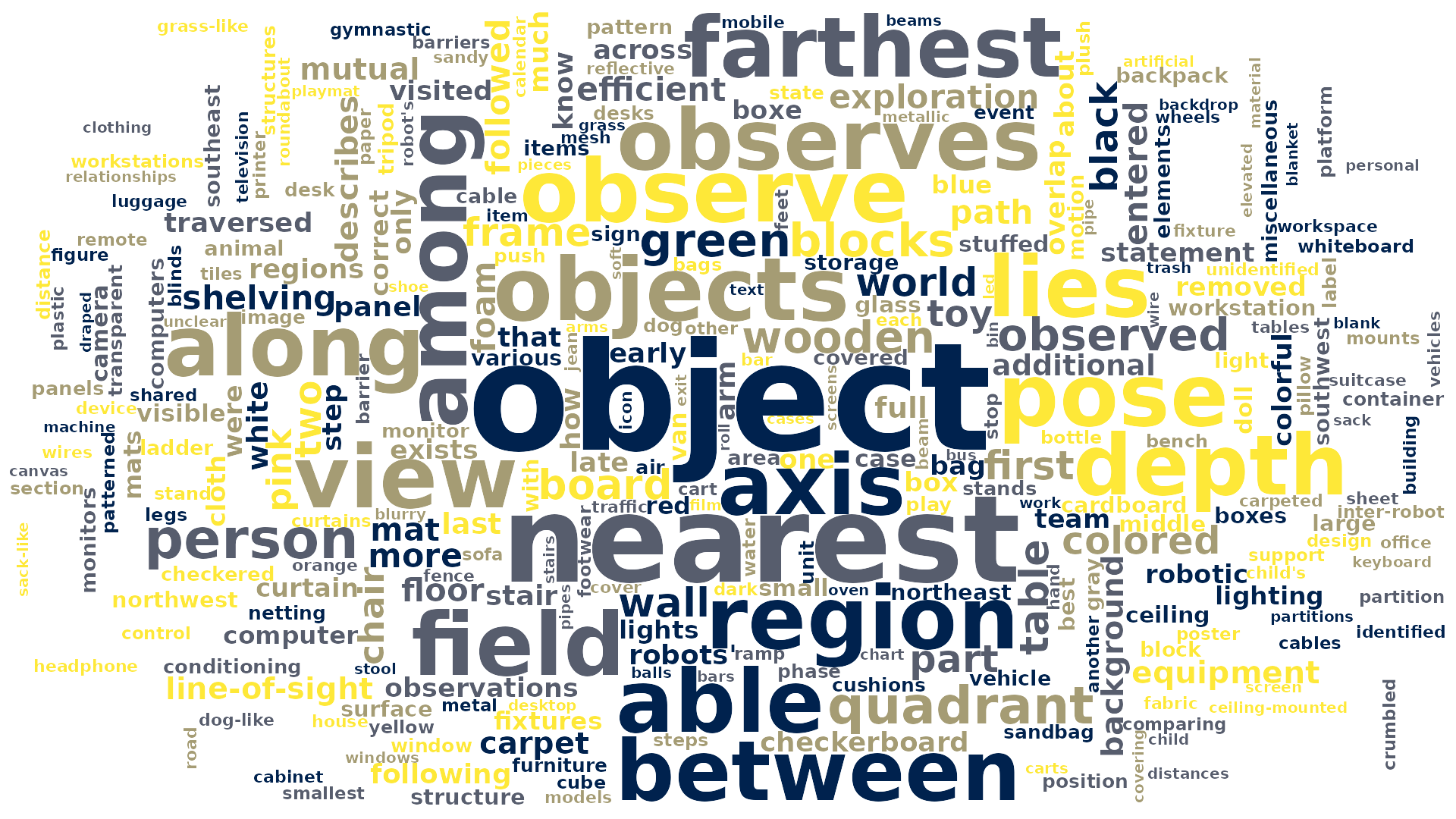}
        \caption{Real-world set}
        \label{fig:image3_realworld}
    \end{subfigure}
    \caption{An overview of the word clouds of the QAs from (a) Habitat, (b) iGibson, and (c) Real-world lab.}
    \label{fig:wordcloud}
\end{figure*}

\begin{figure*}[t!]
    \centering
    \begin{subfigure}{0.49\textwidth}
        \centering
        \includegraphics[width=\linewidth]{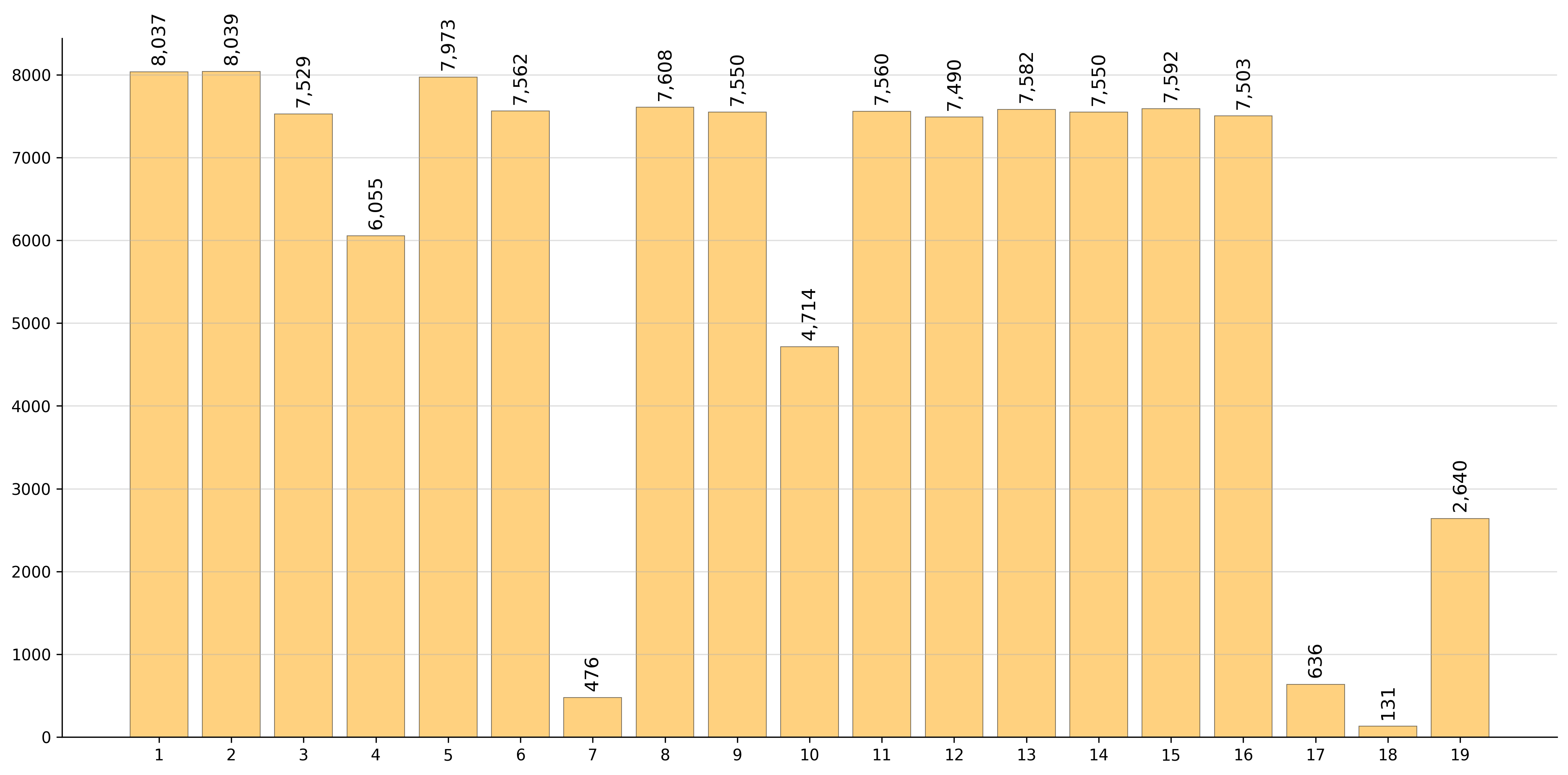}
        \caption{QA statistics.}
        \label{fig:image1_qa_statistics}
    \end{subfigure}
    \hfill
    \begin{subfigure}{0.49\textwidth}
        \centering
        \includegraphics[width=\linewidth]{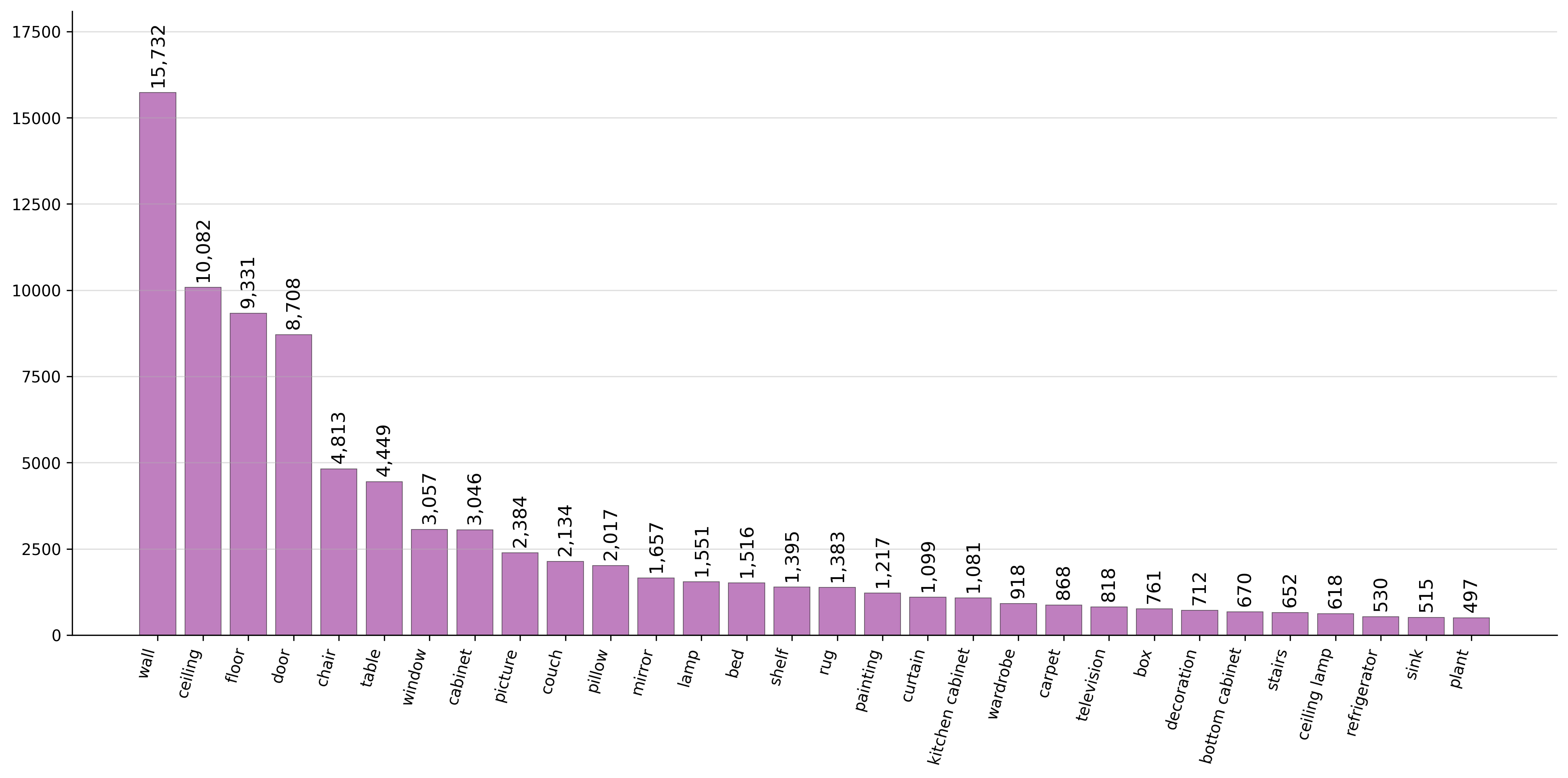}
        \caption{Object statistics from metadata.}
        \label{fig:image2_object_statistics}
    \end{subfigure}
    \caption{An overview of the statistics of (a) the number of QAs and (b) the number of objects.}
    \label{fig:statistics}
\end{figure*}

We analyze the linguistic and semantic diversity of \textbf{CoopSR} in Figures~\ref{fig:wordcloud} and~\ref{fig:statistics}. Figure~\ref{fig:wordcloud} shows word clouds for the Habitat~\cite{puig2023habitat}, iGibson~\cite{li2021igibson}, and real-world lab subsets. Across domains, frequent terms such as ``timestep'', ``observations'', ``robot'', ``object'', ``team'', and ``between'' indicate a shared focus on temporally grounded multi-robot egocentric reasoning. Habitat emphasizes spatial relations with terms such as ``left'', ``right'', ``distance'', and ``closest'', whereas iGibson~\cite{li2021igibson} highlights interaction-rich household scenarios through words such as ``chair'', ``cabinet'', ``table'', ``action'', and ``pushing''. The real-world subset contains prominent terms such as ``pose'', ``nearest'', ``farthest'', ``field'', ``region'', and ``depth'', reflecting visibility-, distance-, and pose-aware reasoning.

Overall, Figure~\ref{fig:wordcloud} suggests that \textbf{CoopSR} spans diverse simulation and real-world language patterns while consistently targeting cooperative spatial, temporal, visibility, and action reasoning.
Figure~5 provides quantitative statistics of the QA and object distributions. Figure~\ref{fig:image1_qa_statistics} reports the number of QA samples for the $19$ QA types. 
Most QA categories contain approximately $7K$--$8K$ samples, indicating broad and relatively balanced coverage across the benchmark taxonomy. A few categories contain fewer samples because they correspond to more constrained events or higher-level cooperative reasoning cases that require specific scene configurations, such as rare interaction events, team-level belief updates, or task-assignment conditions. This distribution ensures that the benchmark contains sufficient supervision for common spatial and temporal reasoning tasks while still including challenging long-tail cooperative reasoning cases.

Figure~\ref{fig:image2_object_statistics} summarizes the object-frequency distribution extracted from metadata. The object distribution is naturally long-tailed: structural and frequently visible categories such as walls, chairs, floors, doors, and cabinets appear most often, while smaller or scene-specific objects such as pillows, lamps, mirrors, sinks, flowers, televisions, and decorative objects occur less frequently. This long-tail distribution reflects realistic indoor environments, where common background structures dominate visual observations while many task-relevant objects appear sparsely. Such diversity is important for evaluating whether MLLMs can reason beyond frequent object categories and generalize to less common objects under partial, egocentric, and multi-view observations.

Together, Figures~\ref{fig:wordcloud} and~\ref{fig:statistics} demonstrate that \textbf{CoopSR} provides both linguistic diversity and semantic richness. The benchmark is not limited to simple object recognition or single-view spatial questions; instead, it contains diverse question formulations, multiple reasoning types, varied object categories, and both simulated and real-world domains. These statistics support the effectiveness of \textbf{CoopSR} as a comprehensive benchmark for evaluating cooperative multi-robot egocentric spatial reasoning.

\subsection{More Ablation Analysis of our SG-CoR Approach on Hyperparameters}

\begin{table*}[t]
\centering
\caption{Ablation studies of SGI-MRF and GMIPSD.}
\label{tab:abl_spi_GMIPSD}

\begin{subtable}[t]{0.49\textwidth}
\centering
\caption{Ablation study of the fusion strategy used in SGI-MRF.}
\label{tab:abl_view_fusion}
\resizebox{\linewidth}{!}{
\begin{tabular}{lllllllll}
\toprule
\multirow{2}{*}{\textbf{Method}} 
& \multicolumn{4}{c}{\textbf{Habitat~\cite{puig2023habitat}}} 
& \multicolumn{4}{c}{\textbf{iGibson~\cite{li2021igibson}}} \\
& \textbf{N=2} & \textbf{N=3} & \textbf{N=4} & \textbf{AVG}
& \textbf{N=2} & \textbf{N=3} & \textbf{N=4} & \textbf{AVG} \\
\midrule
Concatenation  & 72.17 & 60.55 & 56.66 & 62.89 & 60.66 & 62.23 & 62.12 & 61.67 \\
Addition       & 65.61 & 58.61 & 54.56 & 59.43 & 46.75 & 51.64 & 51.95 & 50.14 \\
Multiplication & 69.24 & 59.87 & 56.76 & 61.77 & 51.51 & 57.73 & 55.08 & 54.76 \\
SelfAttention  &  75.65&   66.80 &  65.00 & 68.99 &  64.78 &  67.48 &  66.46&  66.23   \\
\midrule
Ours & \textbf{78.26} & \textbf{67.34} & \textbf{66.60} & \textbf{70.55} & \textbf{72.74} & \textbf{69.92} & \textbf{69.85} & \textbf{70.82} \\
\bottomrule
\end{tabular}}
\end{subtable}
\hfill
\begin{subtable}[t]{0.49\textwidth}
\centering
\caption{Ablation of the loss weight for $L_{\mathrm{distill}}$ ranging from $[0.1, 0.7]$.}
\label{tab:abl_distill}
\resizebox{\linewidth}{!}{
\begin{tabular}{lllllllll}
\toprule
\multirow{2}{*}{\textbf{$L_{\mathrm{distill}}$ in GMIPSD}} 
& \multicolumn{4}{c}{\textbf{Habitat}~\cite{puig2023habitat}} 
& \multicolumn{4}{c}{\textbf{iGibson}~\cite{li2021igibson}} \\
& \textbf{N=2} & \textbf{N=3} & \textbf{N=4} & \textbf{AVG}
& \textbf{N=2} & \textbf{N=3} & \textbf{N=4} & \textbf{AVG} \\
\midrule
0.1 & 77.63 & 67.62 & 64.86 & 69.84 & 68.25 & 68.79 & 67.33 & 68.11 \\
0.3& \textbf{78.26} & 67.34 & 66.60 & \textbf{70.55} & \textbf{72.74} & 69.92 & 69.85 & 70.82 \\
0.5 & 72.19 & \textbf{71.13} & \textbf{70.81} & 70.54 & 72.19 & 71.13 & \textbf{70.81} & 71.37 \\
0.7 & 78.12 & 66.84 & 65.76 & 70.05 & 72.46 & \textbf{72.63} & 69.50 &  \textbf{71.49} \\
\bottomrule
\end{tabular}}
\end{subtable}

\end{table*}
We further analyze the sensitivity of \textbf{SG-CoR} to three important hyperparameters: the distillation loss weight in GMIPSD, the number of learnable prompts, and the number of dense frames retained during CLIP-based similarity pre-selection. The results are reported on both Habitat~\cite{puig2023habitat} and iGibson~\cite{li2021igibson} across different robot-team sizes.
\begin{table*}[t]
\centering
\caption{Ablation studies of dense frame selection and learnable prompts.}
\label{tab:abl_dense_prompts}

\begin{subtable}[t]{0.49\textwidth}
\centering
\caption{Ablation for dense frame selection during the CLIP-based similarity pre-selection.}
\label{tab:abl_dense}
\resizebox{\linewidth}{!}{
\begin{tabular}{lllllllll}
\toprule
\multirow{2}{*}{\textbf{\#Dense Frames}} 
& \multicolumn{4}{c}{\textbf{Habitat}~\cite{puig2023habitat}} 
& \multicolumn{4}{c}{\textbf{iGibson}~\cite{li2021igibson}} \\
& \textbf{N=2} & \textbf{N=3} & \textbf{N=4} & \textbf{AVG}
& \textbf{N=2} & \textbf{N=3} & \textbf{N=4} & \textbf{AVG} \\
\midrule
16 & 76.91 & \textbf{67.66} & 65.88 & 69.98 & 69.26 & 69.17 & 67.94 & 68.77 \\
32 & \textbf{78.26} & 67.34 & \textbf{66.60} & \textbf{70.55} & \textbf{72.74} & \textbf{69.92} & \textbf{69.85} & \textbf{70.82} \\
64 & 77.48 & 66.62 & 64.46 & 69.32 & 67.52 & 69.17 & 67.68 & 68.11 \\
\bottomrule
\end{tabular}}
\end{subtable}
\hfill
\begin{subtable}[t]{0.49\textwidth}
\centering
\caption{Ablation of the number of learnable prompts ranging from $[4,10]$.}
\label{tab:num_prompts}
\resizebox{\linewidth}{!}{
\begin{tabular}{lllllllll}
\toprule
\multirow{2}{*}{\textbf{\#Prompts in GMIPSD}} 
& \multicolumn{4}{c}{\textbf{Habitat}~\cite{puig2023habitat}} 
& \multicolumn{4}{c}{\textbf{iGibson}~\cite{li2021igibson}} \\
& \textbf{N=2} & \textbf{N=3} & \textbf{N=4} & \textbf{AVG}
& \textbf{N=2} & \textbf{N=3} & \textbf{N=4} & \textbf{AVG} \\
\midrule
4  & 77.12 & 67.70 & \textbf{66.69} & 70.34 & 69.99 & 69.82 & 67.33 & 69.01 \\
6  & 77.20 & 67.66 & 64.88 & 69.73 & 69.99 & 69.82 & 67.16 & 68.95 \\
8 & \textbf{78.26} & 67.34 & 66.60 & \textbf{70.55} & \textbf{72.74} & \textbf{69.92} & \textbf{69.85} & \textbf{70.82} \\
10 & 77.65 & \textbf{68.05} & 65.21 & 70.11 & 68.07 & 69.82 & 68.55 & 68.80 \\
\bottomrule
\end{tabular}}
\end{subtable}
\end{table*}
Table~\ref{tab:abl_view_fusion} compares different fusion strategies in SGI-MRF. Overall, the proposed fusion strategy consistently achieves the best performance on both Habitat and iGibson, with average accuracies of 70.55\% and 70.82\%, respectively. Compared with simple concatenation, addition, and multiplication, our method provides more effective multi-view feature integration, indicating that directly combining view features is insufficient to fully capture complementary spatial cues. Self-attention performs better than these simple fusion baselines, suggesting that adaptive view-wise feature interaction is beneficial. However, our method further improves over self-attention by 1.56\% on Habitat~\cite{puig2023habitat} and 4.59\% on iGibson~\cite{li2021igibson} in terms of average accuracy, demonstrating its stronger capability in modeling cross-view relationships and selecting informative features across different numbers of observed views.

First, we vary the weight of the prompt-space distillation loss, $\lambda_d$, from $0.1$ to $0.7$ in Table~\ref{tab:abl_distill}. 
The results show that GMIPSD is effective across a broad range of weights. On Habitat~\cite{puig2023habitat}, the average accuracy remains stable, ranging from $69.84\%$ to $70.55\%$, indicating that the model is not overly sensitive to the exact distillation weight. 
On iGibson~\cite{li2021igibson}, increasing $\lambda_d$ generally improves performance, with the average accuracy increasing from $68.11\%$ at $\lambda_d=0.1$ to $71.49\%$ at $\lambda_d=0.7$. This suggests that stronger geometry- and motion-informed prompt supervision is particularly beneficial in iGibson~\cite{li2021igibson}, where scenes contain more complex interactive dynamics. 
However, overly large weights may slightly reduce performance for some team-size settings, implying that the distillation objective should complement, rather than dominate, the task supervision.

Second, we study the number of learnable prompts in GMIPSD as shown in Table~\ref{tab:num_prompts}.
The performance remains relatively stable when using $4$ to $10$ prompts, demonstrating the robustness of the prompt-space distillation design. 
Using $8$ prompts achieves a strong balance across both environments, reaching $69.99\%$ average accuracy on Habitat~\cite{puig2023habitat} and $70.07\%$ on iGibson~\cite{li2021igibson}. 
Increasing the number of prompts to $10$ slightly improves Habitat performance to $70.11\%$, but decreases iGibson~\cite{li2021igibson} accuracy to $68.80\%$. 
This indicates that a larger prompt capacity does not necessarily lead to better generalization, likely because excessive prompts introduce redundant latent factors or overfit to environment-specific patterns. Therefore, a moderate number of prompts provides sufficient capacity to encode distilled spatial priors while maintaining stable generalization.

Third, we evaluate the number of dense frames used during CLIP-based similarity pre-selection, as shown in Table~\ref{tab:abl_dense}. Increasing the dense-frame budget from $16$ to $32$ improves the average accuracy from $69.98\%$ to $70.55\%$ on Habitat~\cite{puig2023habitat} and from $68.77\%$ to $70.82\%$ on iGibson~\cite{li2021igibson}.
This confirms that retaining more candidate frames before spectral refinement helps preserve useful temporal and semantic evidence. However, further increasing the budget to $64$ does not yield additional gains, leading to $69.99\%$ on Habitat and $70.07\%$ on iGibson~\cite{li2021igibson}. 
This suggests that a moderate candidate set is sufficient for capturing informative moments, while too many dense frames may introduce redundant or noisy visual evidence.

Overall, these ablations show that hyperparameter sensitivity of \textbf{SG-CoR} is acceptable. GMIPSD benefits from a properly weighted distillation objective and a moderate number of learnable prompts, while the frame sampler performs best when the CLIP-based candidate set balances evidence coverage and redundancy reduction. 
These findings further support the design of \textbf{SG-CoR}, where geometry- and motion-informed prompt distillation and spectral frame selection jointly improve cooperative multi-robot spatial reasoning.

\subsection{More Qualitative Analysis}
\begin{table}[t!]
\caption{Comparison to text-only baseline on \textbf{CoopSR} benchmark.}
\label{tab:textonly}
\centering
\resizebox{\columnwidth}{!}{
\begin{tabular}{l|lll|l|lll|l}
\toprule
\textbf{Datasets}  & \multicolumn{4}{c|}{\textbf{Habitat}~\cite{puig2023habitat}}                                                               & \multicolumn{4}{c}{\textbf{iGibson}~\cite{li2021igibson}}                                  \\
\textbf{Models}                 & \textbf{N=2} & \textbf{N=3} & \textbf{N=4} &\textbf{AVG} &\textbf{N=2} & \textbf{N=3} & \textbf{N=4} & \textbf{AVG} \\
\midrule
Text-Only Baseline &23.51 & 27.27 & 29.97 &29.47 & 26.58 & 30.31& 31.30& 26.97 \\
SFT (Qwen2.5-VL-7B-Instruct~\cite{bai2025qwen25vl})  & 75.04& 63.82    &  61.77  &  66.68 & 63.40   & 63.92& 63.68& 63.67   \\
\midrule
\textbf{SG-CoR} (Qwen2.5-VL-7B-Instruct~\cite{bai2025qwen25vl})  & \textbf{78.26} & \textbf{67.34} & \textbf{66.60} & \textbf{70.55} & \textbf{72.74} & \textbf{69.92} & \textbf{69.85} & \textbf{70.82} \\
\bottomrule
\end{tabular}}
\vskip-3ex
\end{table}
\begin{table}[t!]
\caption{Comparison of our approach and the counterpart under robot Dropout setting.}
\label{tab:missing_view}
\centering
\resizebox{\columnwidth}{!}{
\begin{tabular}{l|llll|l|llll|l}
\toprule
\textbf{Datasets}  & \multicolumn{5}{c|}{\textbf{Habitat}~\cite{puig2023habitat}}                                                               & \multicolumn{5}{c}{\textbf{iGibson}~\cite{li2021igibson}}                                  \\
\textbf{Models}                 & \textbf{T=1} & \textbf{T=2} & \textbf{T=3} & \textbf{T=4} &\textbf{AVG} & \textbf{T=1} &\textbf{T=2} & \textbf{T=3} & \textbf{T=4} & \textbf{AVG} \\
\midrule

Robot Dropout on SG-CoR (Qwen2.5-VL-7B-Instruct~\cite{bai2025qwen25vl}) &   31.69  &  58.29  &  44.10  &  59.74  & 53.29   &  48.89  &  52.38  &  43.31 &   48.64  & 48.35  \\
\midrule
\textbf{SG-CoR} (Qwen2.5-VL-7B-Instruct~\cite{bai2025qwen25vl}) & \textbf{55.36} & \textbf{78.72} & \textbf{59.36} & \textbf{74.94}  & \textbf{70.55} & \textbf{74.44} & \textbf{71.88} & \textbf{58.36}  & \textbf{74.63} & \textbf{70.82} \\
\bottomrule
\end{tabular}}
\vskip-3ex
\end{table}

\begin{figure}[t!]
\includegraphics[width=\linewidth]{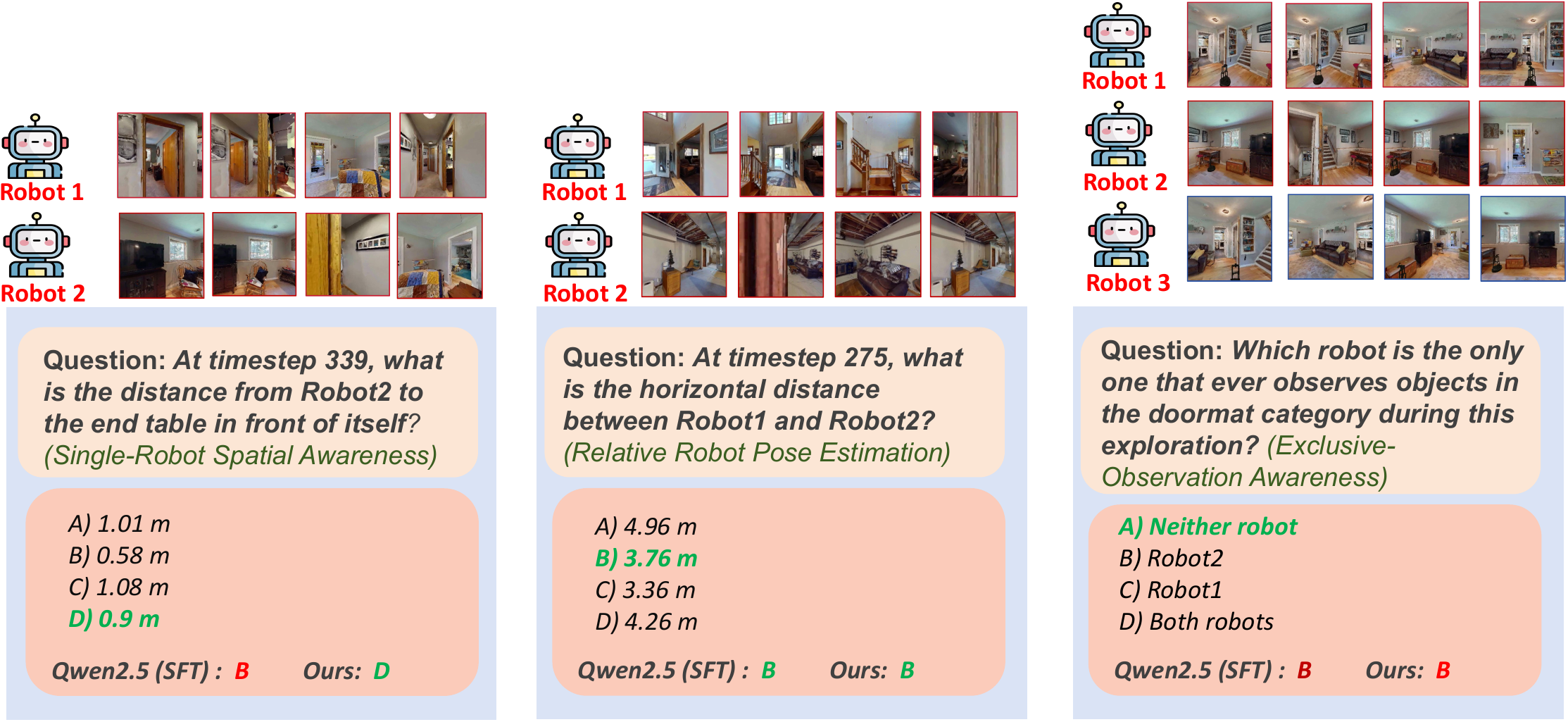}
    \caption{Overview of qualitative results of our approach and SFT Qwen2.5-VL-7B~\cite{hui2024qwen2}.}
    \label{fig:overview_qualitative_results1}
\end{figure}

\begin{figure}[t!]
\includegraphics[width=\linewidth]{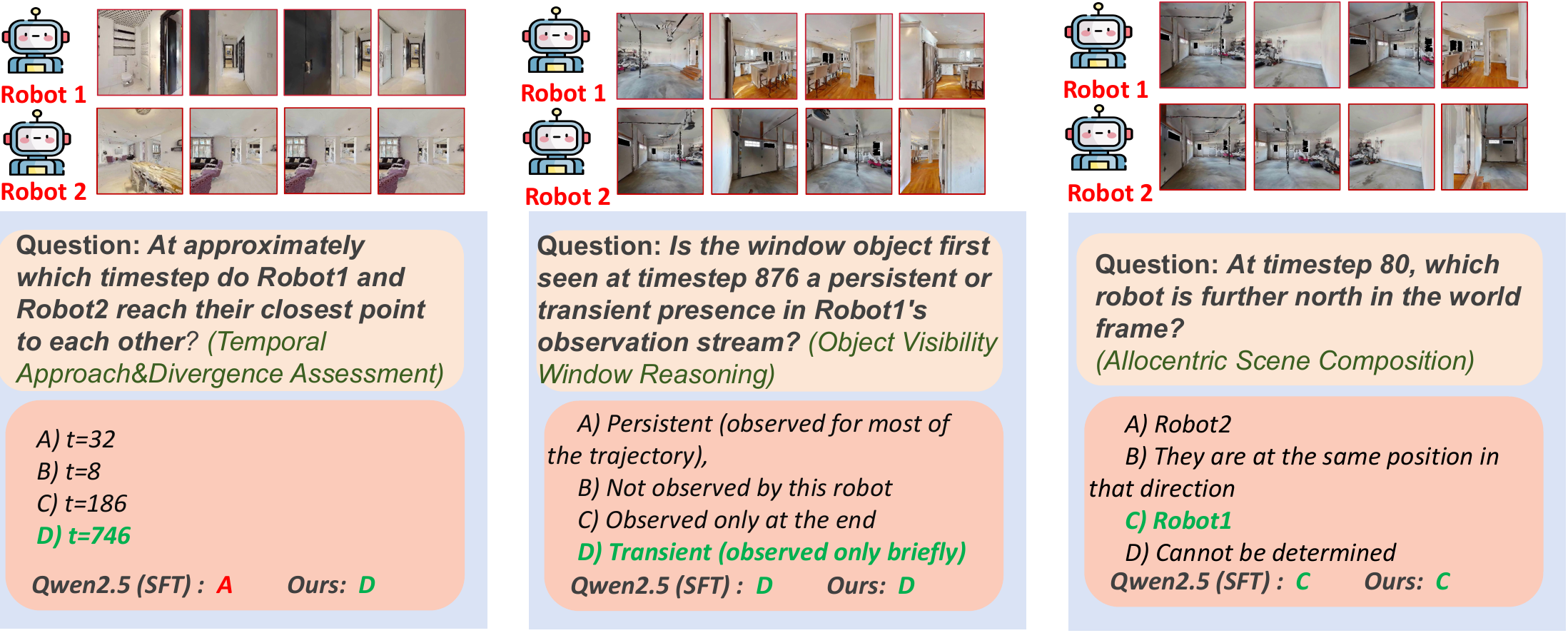}
    \caption{Overview of qualitative results of our approach and SFT Qwen2.5-VL-7B~\cite{hui2024qwen2}.}
\label{fig:overview_qualitative_results2}
\end{figure}

\begin{figure}[t!]
\includegraphics[width=\linewidth]{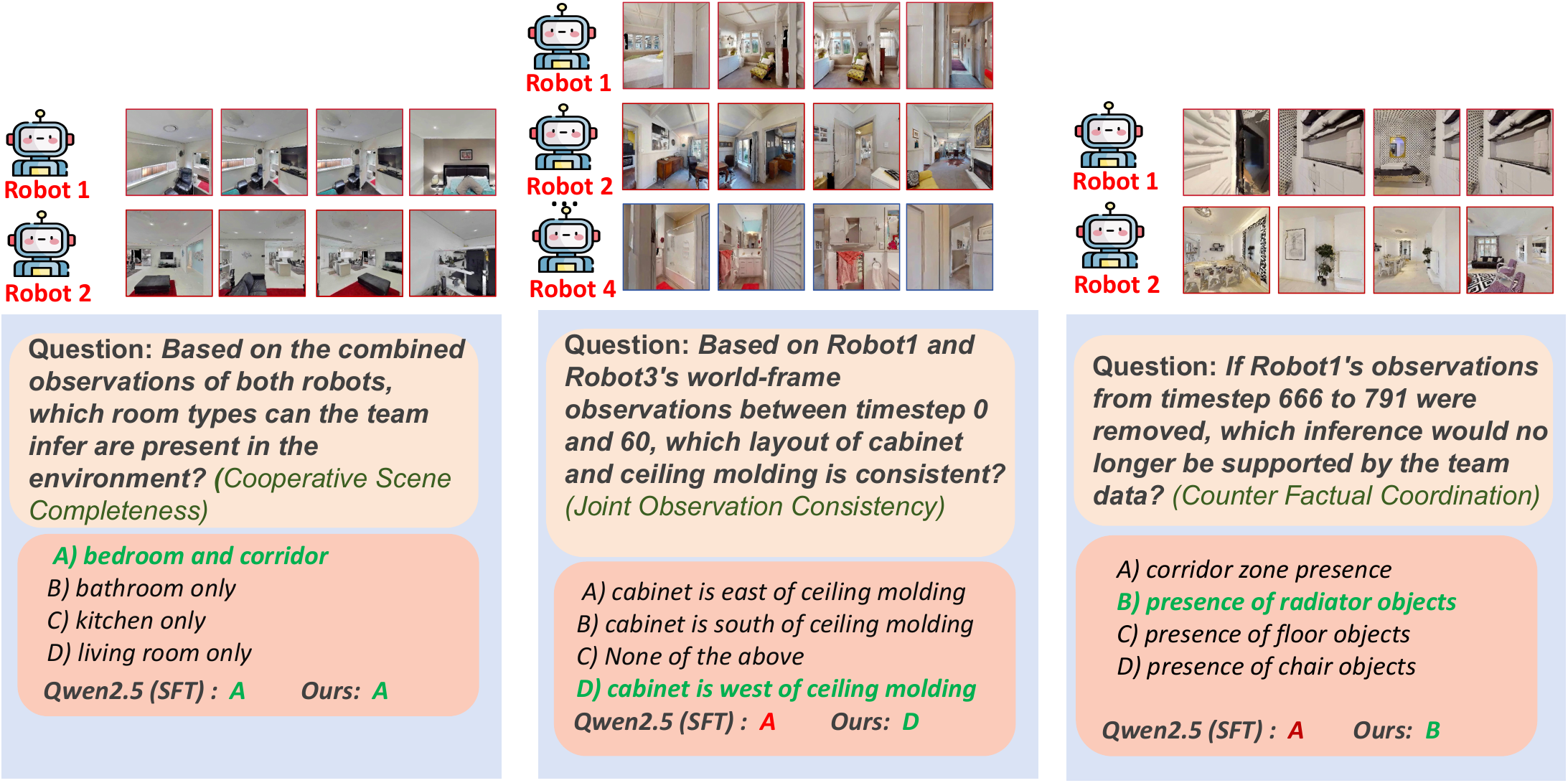}
    \caption{Overview of qualitative results of our approach and SFT Qwen2.5-VL-7B~\cite{hui2024qwen2}.}
\label{fig:overview_qualitative_results3}
\end{figure}

\begin{figure}[t!]
\includegraphics[width=\linewidth]{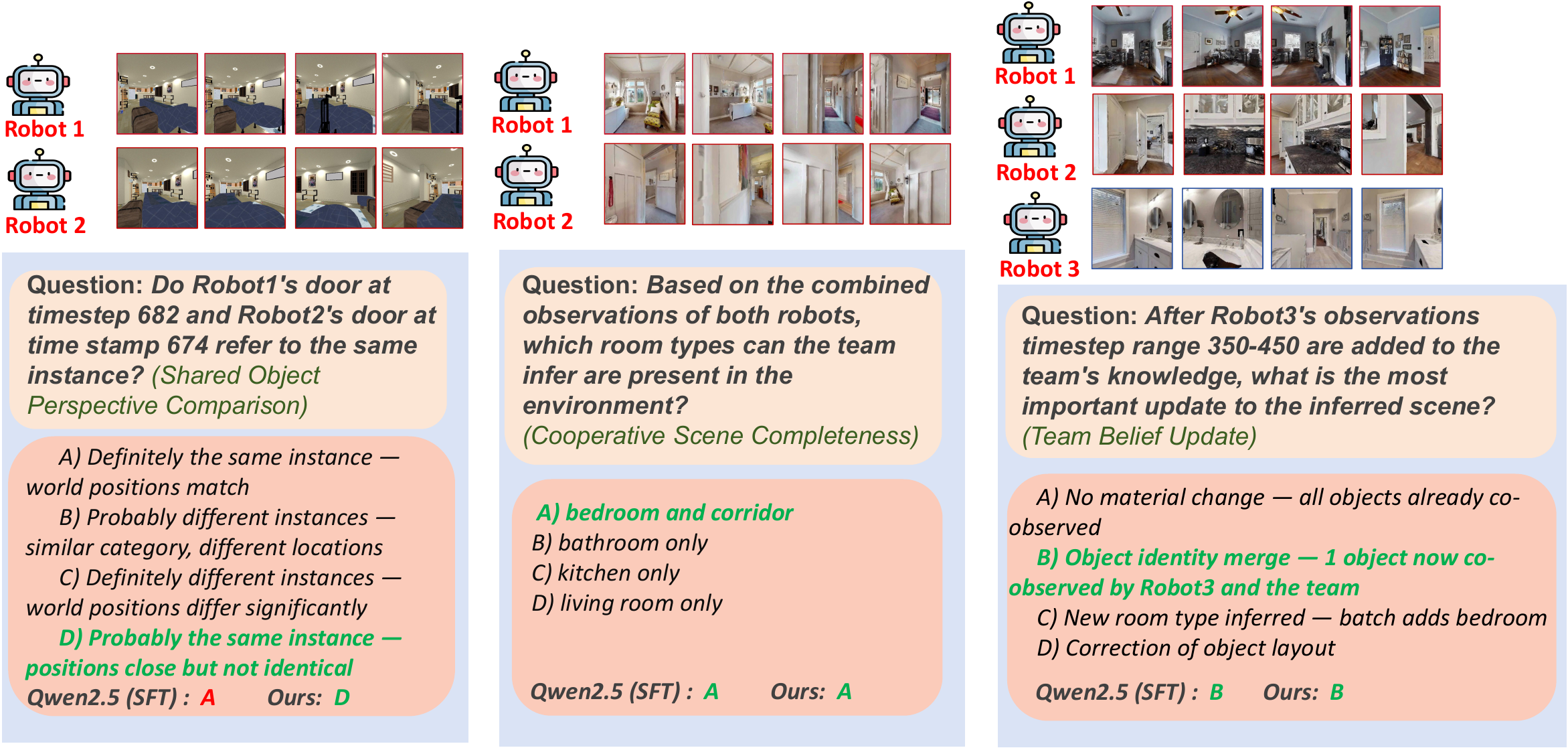}
    \caption{Overview of qualitative results of our approach and SFT Qwen2.5-VL-7B~\cite{hui2024qwen2}.}
\label{fig:overview_qualitative_results4}
\end{figure}

\begin{figure}[t!]
\includegraphics[width=\linewidth]{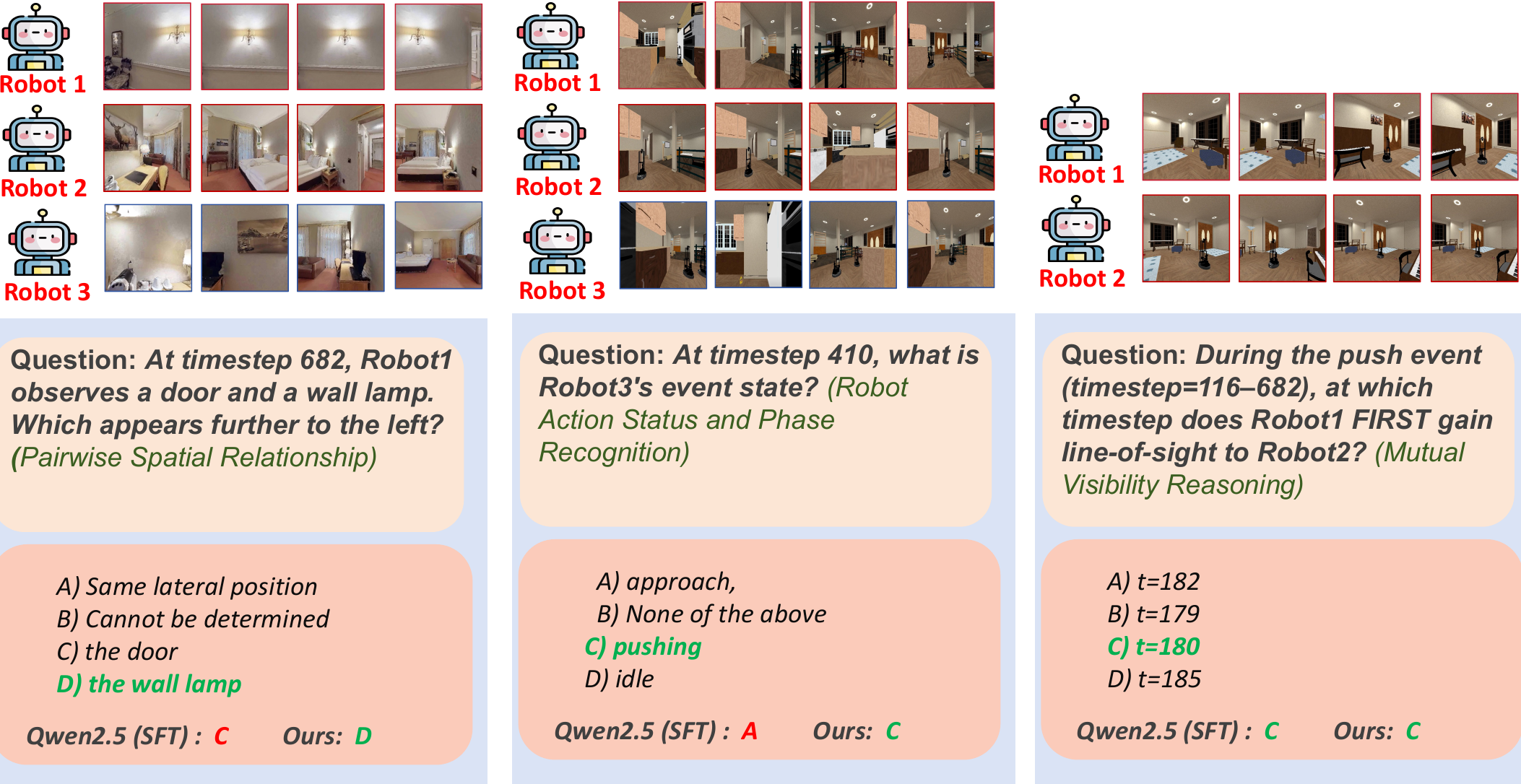}
    \caption{Overview of qualitative results of our approach and SFT Qwen2.5-VL-7B~\cite{hui2024qwen2}.}
\label{fig:overview_qualitative_results5}
\end{figure}

\begin{figure}[t!]
\includegraphics[width=\linewidth]{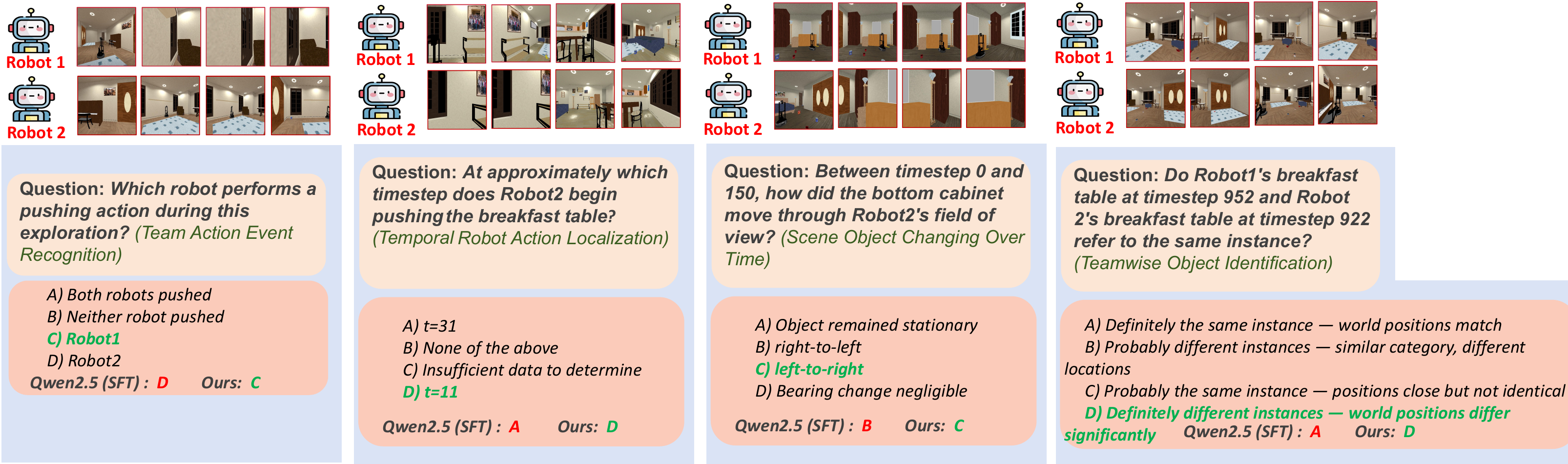}
    \caption{Overview of qualitative results of our approach and SFT Qwen2.5-VL-7B~\cite{hui2024qwen2}.}
\label{fig:overview_qualitative_results6}
\end{figure}

Figure~\ref{fig:overview_qualitative_results1} to Figure~\ref{fig:overview_qualitative_results6} compare \textbf{SG-CoR} with the SFT (Qwen2.5-VL-7B~\cite{bai2025qwen25vl}) baseline on representative \textbf{CoopSR} examples. \textbf{SG-CoR} is more accurate on tasks requiring cross-robot view integration, temporal event localization, mutual visibility reasoning, and teamwise object association. The SFT baseline often confuses robot identities, action phases, or visually similar objects across views, suggesting that standard fine-tuning mainly adapts the backbone to the QA format but does not provide explicit cooperative inductive biases.

These examples highlight the novelty of \textbf{SG-CoR}: rather than treating multi-robot videos as concatenated visual tokens, it introduces dynamics-aware frame sampling to retain informative temporal evidence, geometry- and motion-informed fusion to align observations under changing egocentric viewpoints, and prompt-space distillation to transfer pose-level geometric priors without requiring poses at test time. As a result, \textbf{SG-CoR} can build a more consistent team-level scene representation, enabling robust reasoning over partial observations, inter-robot relations, and dynamic cooperative events.

\subsection{Comparison with Text-Only Baseline}
We include a text-only baseline to evaluate whether \textbf{CoopSR} can be solved using linguistic priors or dataset biases without visual observations. As shown in Table~\ref{tab:textonly}, the text-only baseline obtains much lower accuracy than vision-based models, with only 29.47\% on Habitat and 26.97\% on iGibson, compared with 66.68\% and 63.67\% for the SFT baseline. This large gap demonstrates that \textbf{CoopSR} requires grounding in synchronized multi-robot egocentric videos, while the further gains of \textbf{SG-CoR} show the benefit of explicit cooperative spatial reasoning.

\subsection{Importance of Multi-Robot Cooperative Reasoning and Experiments of Missing Robot}

To verify that \textbf{CoopSR} genuinely requires cooperative reasoning across multiple robots rather than relying on a single informative viewpoint, we conduct a robot-dropout analysis, where one robot stream is randomly removed during inference. Results are shown in Table~\ref{tab:missing_view}. Removing one robot consistently degrades performance across all reasoning tiers and datasets. On Habitat, the average accuracy drops from 70.55\% to 53.29\%, while on iGibson~\cite{li2021igibson} it decreases from 70.82\% to 48.35\%.

The degradation is especially clear on higher-level cooperative reasoning tasks, where models must integrate complementary observations to infer scene-level structure, mutual visibility, temporal interactions, and team belief updates. These results show that \textbf{CoopSR} cannot be reliably solved from a reduced set of robot views alone. Missing one robot removes critical evidence for resolving occlusions, cross-view object grounding, and inter-robot spatial relations. The substantial gap between Robot Dropout and full \textbf{SG-CoR} confirms that the benchmark evaluates genuine multi-robot cooperative spatial reasoning rather than independent single-robot perception.

\subsection{Additional Communication Baselines and Failure Analysis}
\label{app:communication_failure}

\paragraph{Decentralized language-communication baselines.}
We evaluate two additional baselines on the Habitat test set,
with all methods using the Qwen2.5-VL-7B backbone.
These baselines process each robot's video independently
before combining information through language or voting.
In \emph{Structured Report Aggregator}, each robot converts
its observations into a robot-indexed textual report, and
an LLM aggregates the reports to answer the question.
In \emph{Confidence-Weighted Vote}, each robot independently
predicts an answer and confidence score, and the final answer
is selected by confidence-weighted voting.

\begin{table}[t]
    \centering
    \caption{Comparison of centralized and decentralized
    approaches on Habitat, measured by accuracy (\%).
    All methods use Qwen2.5-VL-7B.
    T1--T4 denote reasoning tiers, and $N$ denotes team size.
    Decentralized methods combine locally generated reports
    or predictions rather than jointly processing all videos.
    Best results are in bold.}
    \label{tab:decentralized_habitat}
    \small
    \setlength{\tabcolsep}{4pt}
    \resizebox{\linewidth}{!}{%
    \begin{tabular}{llrrrrrrr}
        \toprule
        & & \multicolumn{4}{c}{Reasoning tier}
          & \multicolumn{3}{c}{Team size} \\
        \cmidrule(lr){3-6} \cmidrule(lr){7-9}
        Setting & Method
        & T1 & T2 & T3 & T4
        & $N=2$ & $N=3$ & $N=4$ \\
        \midrule
        Centralized & Zero-shot
        & 23.57 & 39.61 & 28.40 & 34.35
        & 34.66 & 34.99 & 31.03 \\
        Decentralized & Confidence-Weighted Vote
        & 24.44 & 38.98 & 26.79 & 27.89
        & 31.79 & 33.94 & 25.78 \\
        Decentralized & Structured Report Aggregator
        & 31.11 & 48.02 & 30.36 & 47.81
        & 38.97 & 44.34 & 42.67 \\
        \midrule
        Centralized & SFT
        & 46.67 & 76.30 & 53.02 & 72.53
        & 75.04 & 63.82 & 61.77 \\
        Centralized & EgoRAG
        & 48.31 & 74.93 & 51.80 & 71.36
        & 73.51 & 63.61 & 60.37 \\
        Centralized & TSPO
        & 37.97 & 75.80 & 46.68 & 68.55
        & 69.28 & 61.44 & 57.94 \\
        Centralized & \textbf{SG-CoR}
        & \textbf{55.36} & \textbf{78.72}
        & \textbf{59.36} & \textbf{74.94}
        & \textbf{78.26} & \textbf{67.34}
        & \textbf{66.60} \\
        \bottomrule
    \end{tabular}%
    }
\end{table}

\paragraph{Quantitative comparison.}
As shown in Table~\ref{tab:decentralized_habitat},
Structured Report Aggregator outperforms both
Confidence-Weighted Vote and the centralized zero-shot
baseline across all reported reasoning tiers and team sizes.
These results suggest that exchanging structured textual
evidence can support cooperative reasoning more effectively
than aggregating independent answer predictions.
Nevertheless, \textbf{SG-CoR} exceeds Structured Report Aggregator
by 27.13\%--30.70\% on T2--T4 and
23.00\%--39.29\% across team sizes.

One possible limitation of report-based aggregation is that
converting each stream into text before cross-robot comparison
may discard fine-grained visual, temporal, or geometric
evidence. Such information is relevant to relative pose,
mutual visibility, cross-view object identity, observation
consistency, and team-belief reasoning.
Language communication therefore offers a complementary
approach to cooperative reasoning, while the results motivate
further investigation of representations that preserve
cross-view evidence. These comparisons evaluate the reported
systems; isolating the effect of communication format would
require matched training and inference budgets.

\paragraph{Preliminary failure-case analysis.}
Qualitative inspection of \textbf{SG-CoR} errors identifies three
recurring limitations.
\emph{First, insufficient visual evidence:}
severe occlusion or small, distant objects can make the
relevant evidence difficult to recover across all robot views.
This motivates occlusion-aware evidence retrieval and
uncertainty-based abstention when the available observations
remain insufficient.
\emph{Second, incomplete temporal coverage:}
the fixed frame budget can miss brief events or precise
action transitions when decisive evidence occurs between
selected frames, motivating adaptive temporal sampling.
\emph{Third, cross-view association errors:}
visually similar robot streams and contradictory observations
can still lead to agent-identity confusion or incorrect
object associations, motivating stronger identity-preserving
cross-view alignment.
These observations provide qualitative directions for
improvement; they do not constitute a quantitative breakdown
of failure frequencies.

\subsection{Contributions of Visual, Role, and Privileged Pose Cues}
\label{app:cue_ablation}

\textbf{SG-CoR} performs inference using robot-indexed egocentric RGB
videos and the textual query, with the active robot count
provided as structural metadata. Visual features provide
appearance and spatial evidence, spectral features summarize
temporal changes, and learned role embeddings encode
episode-local robot identities. Motion and pose information
provide privileged supervision during training through GMIPSD
and are not required at inference.

To examine these contributions, we evaluate variants that
remove the role token, visual token, or training-time pose
token. Table~\ref{tab:cue_ablation} reports results using the
Qwen2.5-VL-7B backbone on Habitat and iGibson.
The pose-token ablation concerns privileged training
information, rather than a test-time pose input.

\begin{table}[t!]
    \centering
    \caption{Ablation of role, visual, and training-time pose
    tokens using Qwen2.5-VL-7B. Accuracy (\%) is reported
    across team sizes on Habitat and iGibson.
    AVG denotes accuracy over all evaluated QA samples.
    Best results are in bold.}
    \label{tab:cue_ablation}
    \small
    \setlength{\tabcolsep}{4pt}
    \resizebox{\linewidth}{!}{%
    \begin{tabular}{lrrrrrrrr}
        \toprule
        & \multicolumn{4}{c}{Habitat}
        & \multicolumn{4}{c}{iGibson} \\
        \cmidrule(lr){2-5} \cmidrule(lr){6-9}
        Method
        & $N=2$ & $N=3$ & $N=4$ & AVG
        & $N=2$ & $N=3$ & $N=4$ & AVG \\
        \midrule
        w/o role token
        & 72.55 & 62.98 & 59.32 & 64.75
        & 60.38 & 65.14 & 63.16 & 62.88 \\
        w/o visual token
        & 33.37 & 47.42 & 48.24 & 43.28
        & 35.68 & 45.36 & 45.70 & 42.28 \\
        w/o pose token (training)
        & 72.15 & 63.06 & 59.42 & 64.69
        & 59.93 & 64.85 & 62.47 & 62.40 \\
        \midrule
        \textbf{SG-CoR}
        & \textbf{78.26} & \textbf{67.34}
        & \textbf{66.60} & \textbf{70.55}
        & \textbf{72.74} & \textbf{69.92}
        & \textbf{69.85} & \textbf{70.82} \\
        \bottomrule
    \end{tabular}%
    }
\end{table}
\begin{table}[t!]
    \centering
    \caption{Evaluation on seen scenarios using the
    Qwen2.5-VL-Instruct backbone. Three of the fifteen
    explorations per scenario are reserved for testing.
    Accuracy (\%) is reported across robot team sizes,
    and AVG denotes accuracy over all evaluated QA samples.
    Best results are in bold.}
    \label{tab:seen_scenarios}
    \small
    \setlength{\tabcolsep}{4pt}
    \resizebox{\linewidth}{!}{%
    \begin{tabular}{lrrrrrrrr}
        \toprule
        & \multicolumn{4}{c}{Habitat}
        & \multicolumn{4}{c}{iGibson} \\
        \cmidrule(lr){2-5} \cmidrule(lr){6-9}
        Method
        & $N=2$ & $N=3$ & $N=4$ & AVG
        & $N=2$ & $N=3$ & $N=4$ & AVG \\
        \midrule
        SFT (Qwen2.5-VL-Instruct)
        & 61.83 & 61.03 & 55.69 & 59.43
        & 53.42 & 67.61 & 53.95 & 58.18 \\
        \textbf{SG-CoR (Qwen2.5-VL-Instruct)}
        & \textbf{76.08} & \textbf{66.47}
        & \textbf{63.46} & \textbf{68.46}
        & \textbf{70.07} & \textbf{73.38}
        & \textbf{62.34} & \textbf{68.41} \\
        \bottomrule
    \end{tabular}%
    }
\end{table}
\noindent\textbf{Visual evidence.}
Removing the visual token produces the largest decrease
among the evaluated variants: overall accuracy falls by
27.27\% on Habitat and 28.54\% on iGibson. This highlights the importance of direct
visual evidence for cooperative spatial reasoning.
Together with the spectral-energy and Fourier-embedding
ablations in Table~\ref{tab:abl}, these results support the complementary
contributions of visual and spectral features.

\noindent\textbf{Robot identity.}
Removing the role token reduces overall accuracy by
5.80\% on Habitat and 7.94\% on iGibson. These decreases support the value of
explicit robot-identity cues during multi-robot fusion,
beyond organizing the input streams by robot index.

\noindent\textbf{Privileged training information.}
Removing the training-time pose token reduces overall
accuracy by 5.86\% on Habitat and
8.42\% on iGibson. These results support
the benefit of privileged pose supervision for inference
without pose measurements. Together with the GMIPSD ablation
in Table~\ref{tab:abl}, they indicate that training-time geometric
information can improve reasoning from egocentric visual
observations.

\subsection{Evaluation on Seen Scenarios}
\label{app:seen_scenarios}

To complement the scene-disjoint evaluation, we additionally
evaluate \textbf{SG-CoR} on held-out explorations within seen scenarios.
For each scenario, three of the fifteen exploration episodes
are reserved for testing. The split therefore operates at
the exploration level: training and testing share scenarios,
while the test set contains held-out exploration episodes.
This setting evaluates reasoning over new observations
within environments encountered during training.

As shown in Table~\ref{tab:seen_scenarios}, \textbf{SG-CoR} outperforms
SFT across both datasets and all evaluated team sizes.
Its overall accuracy reaches $68.46\%$ on Habitat and
$68.41\%$ on iGibson, exceeding SFT by $9.03\%$ and $10.23\%$, respectively. Across individual team
sizes, the improvements range from $5.44\%$ to $14.25\%$ on Habitat and from $5.77\%$ to $16.65\%$ on iGibson. These results show that
\textbf{SG-CoR}'s performance advantage extends to held-out
explorations within seen scenarios, complementing the
evaluation on unseen scenes.


\end{document}